\documentclass{CUP-JNL-DTM}%

\usepackage{graphicx}
\usepackage{multicol,multirow}
\usepackage{amsmath,amssymb,amsfonts}
\usepackage{mathrsfs}
\usepackage[font=small,labelfont=bf]{caption}
\usepackage{amsthm}
\usepackage{rotating}
\usepackage{appendix}
\usepackage{longtable}
\usepackage{array,ragged2e}
\newcolumntype{P}[1]{>{\RaggedRight\arraybackslash}p{#1}}
\usepackage[natbib,style=apa]{biblatex}
\usepackage{ifpdf}
\usepackage[T1]{fontenc}
\usepackage{newtxtext}
\usepackage{newtxmath}
\usepackage{microtype}
\usepackage{textcomp}
\usepackage{xcolor}
\usepackage{lipsum}
\usepackage{float}
\usepackage[colorlinks,allcolors=blue]{hyperref}
\usepackage[finalnew]{trackchanges} 

\theoremstyle{definition}

\numberwithin{equation}{section}

\jname{Environmental Data Science (Accepted for Publication on May 8th, 2024)}
\articletype{Survey Paper}
\jyear{2024}

\begin{document}

\begin{Frontmatter}

\title[Time Series Predictions in Unmonitored Sites: A Survey of Machine Learning Techniques in Water Resources]{Time Series Predictions in Unmonitored Sites: A Survey of Machine Learning Techniques in Water Resources}


\author[1,4]{Jared D. Willard}
\author[2]{Charuleka Varadharajan}
\author[3]{Xiaowei Jia}
\author[4]{Vipin Kumar}

\authormark{Willard et al.}

\address[1]{\orgdiv{Computing Sciences Area}, \orgname{Lawrence Berkeley National Laboratory}, \orgaddress{\city{Berkeley}, \postcode{94720}, \state{CA},  \country{USA}},\email{jwillard@lbl.gov}}

\address[2]{\orgdiv{Earth and Environmental Sciences Area}, \orgname{Lawrence Berkeley National Laboratory}, \orgaddress{\city{Berkeley}, \postcode{94720}, \state{CA},  \country{USA}},\email{jwillard@lbl.gov,cvaradharajan@lbl.gov}}

\address[3]{\orgdiv{Department of Computer Science}, \orgname{University of Pittsburgh}, \orgaddress{\city{Pittsburgh}, \postcode{15260}, \state{PA},  \country{USA}},\email{xiaowei@pitt.edu}}

\address[4]{\orgdiv{Department of Computer Science and Engineering}, \orgname{University of Minnesota Twin Cities}, \orgaddress{\city{Minneapolis}, \postcode{55455}, \state{MN},  \country{USA}},\email{kumar001@umn.edu}}



\abstract{
Prediction of dynamic environmental variables in unmonitored sites remains a long-standing challenge for water resources science. The majority of the world's freshwater resources have inadequate monitoring of critical environmental variables needed for management. Yet, the need to have widespread predictions of hydrological variables such as river flow and water quality has become increasingly urgent due to climate and land use change over the past decades, and their associated impacts on water resources. Modern machine learning methods increasingly outperform their process-based and empirical model counterparts for hydrologic time series prediction with their ability to extract information from large, diverse data sets. We review relevant state-of-the art applications of machine learning for streamflow, water quality, and other water resources prediction and discuss opportunities to improve the use of machine learning with emerging methods for incorporating watershed characteristics \add{and process knowledge} into \add{classical,} deep learning\add{, and transfer learning methodologies} \remove{models} \remove{, transfer learning, and incorporating process knowledge into machine learning models}. The analysis here suggests most prior efforts have been focused on deep \remove{learning} learning frameworks built on many sites for predictions at daily time scales in the United States, but that comparisons between different classes of machine learning methods are few and inadequate. We identify several open questions for time series predictions in unmonitored sites that include incorporating dynamic inputs and site characteristics, mechanistic understanding and spatial context, and explainable AI techniques in modern machine learning frameworks.  
}

\end{Frontmatter}
 \section*{Impact Statement}
This review addresses a gap that different types of ML methods for hydrological time series prediction in unmonitored sites are often not compared in detail and best practices are unclear. We consolidate and synthesize state-of-the-art ML techniques for researchers and water resources management, where the strengths and limitations of different ML techniques are described allowing for a more informed selection of existing ML frameworks and development of new ones. Open questions that require further investigation are highlighted to encourage researchers to address specific issues like training data and input selection, model explainability, and the incorporation of process-based knowledge. 

\section[Introduction]{Introduction}

Environmental data for water resources often does not exist at the appropriate spatiotemporal resolution or coverage for scientific studies or management decisions.  Although advanced sensor networks and remote sensing are generating more environmental data \citep{reichstein_deep_2019,hubbard2020emerging,topp2020research}, the amount of observations available will continue to be inadequate for the foreseeable future, notably for variables that are only measured at a few locations. For example, the United States Geological Survey (USGS) streamflow monitoring network covers less than 1\% of stream reaches in the United States, with monitoring sites declining over time \citep{ahuja2016chemistry,konrad2022network}, and stream coverage is significantly lower in many other parts of the world. Similarly, just over 12,000 of the 185,000 lakes with at least 4 hectares in area in the conterminous US (CONUS)  have at least one lake surface temperature measurement \citep{willard_daily_2022}, and less than 5\% of those have 10 or more days with temperature measurements \citep{read_water_2017}. Since observing key variables at scale is prohibitively costly \citep{caughlan_cost_2001}, models that use existing data and transfer information to unmonitored systems are critical to closing the data gaps. The problem of streamflow and water quality prediction in unmonitored basins \change{, commonly referred to as "PUBs",}{in particular} has been a longstanding area of research in hydrology due to its importance for infrastructure design, energy production, and management of water resources. The need for \change{PUBs}{these predictions} has grown with changing climate, increased frequency and intensity of extreme events,  and widespread human impacts on water resources \citep{bloschl2013runoff,salinas2013comparative,sanchez2023streamflow,guo_regionalization_2020,zounemat2021ensemble}. 


\remove{Typically} \change{a}{A} variety of models --- process-based, machine learning (ML) and statistical models --- have been used to predict key ecosystem variables\remove{, which otherwise would be} \add{. These models can be applied to a few categories of applications where data are }unavailable at the spatial and temporal scales needed for environmental decision-making. 
\add{ The first is based on data completeness, which could occur when (a) a site is not monitored at all; (b) a site is monitored, but the time series has large chunks of missing data or is available for a limited period; (c) a site is monitored, but the time series has sporadic missing data. A second  is based on data resolution when (a) a site is monitored but at lower resolution than desired, or (b) a site is not monitored but data for other covariates are available. In this paper, we define the problem of predictions in unmonitored sites or the 'unmonitored' scenario as specifically the cases where }\change{the input data to the model is available at all time steps but}{ the locations have either no monitoring data at all for the variable of interest, or sufficiently sparse or low-resolution monitoring data where} \change{it is inadequate to build a data-driven model} {it effectively can be considered as an unmonitored site}\add{.  In cases where data need to be gap filled or extended forward or backwards in time, }\remove{where} a model \change{is}{can be} trained on a time period within one site and then predictions are made for new time periods at the same site. This is often referred to as the \protect\textit{monitored} prediction scenario or the \protect\textit{gauged} scenario in streamflow modeling. While temporal predictions in monitored sites are important, spatial extrapolation to \protect\textit{unmonitored} sites is even more crucial, because the vast majority of locations remain unmonitored for many environmental variables\add{ of interest}. 

Traditionally, water resources modeling in unmonitored sites has relied on the regionalization of process-based models. Regionalization techniques relate the parameter values of a model calibrated to the data of a monitored site to the inherent characteristics of the unmonitored site \citep{seibert1999regionalisation,razavi2013streamflow,yang2019transferability}. However, large uncertainties and mixed success have prevented process-based model regionalization from being widely employed in hydrological analysis and design \citep{wagener2006parameter,bastola2008regionalisation,prieto2019flow}. A major issue that makes process-based model calibration and regionalization difficult is the complex relationships between model parameters (e.g. between soil porosity and soil depth in rainfall-runoff models)  \citep{oudin2008spatial,kratzert2019benchmarking}, which leads to the problem of equifinality \citep{beven2001equifinality} where different parameter values or model structures are equally capable of reproducing a similar hydrological outcome.  Additionally, process models require significant amounts of site-specific data collection and computational power for calibration and benchmarking, which is expensive to generate across diverse regions of interest.  


On the other hand, ML models built using data from large-scale monitoring networks do regionalization implicitly without the dependence on expert knowledge, pre-defined hydrological models, and also often without any hydrological knowledge at all. Since ML models have significantly more flexibility in how parameters and connections between parameters are optimized, unlike process-based models where each parameter represents a specific system component or property, issues relevant to equifinality become largely irrelevant \citep{razavi2022coevolution}. In recent years, numerous ML approaches have been explored for environmental variable time series \change{PUBs}{predictions in unmonitored locations} that span a variety of methods and applications in hydrology and water resources engineering. \change{Though m}{M}ost of the ML approaches \add{for predictions in unmonitored regions focus on}\remove{ are developed for} stream flows, \change{ these efforts are}{ but are rapidly } expanding to other variables\change{like river and lake water quality as data collection and modeling continue to advance}{such as soil moisture \protect\citep{fang2018value}, stream temperature \protect\citep{rahmani2021deep,weierbach2022stream}, and lake temperature \citep{willard_daily_2022}}. The ML \remove{PUB} models have continually outperformed common process-based hydrological models in terms of both predictive performance and computational efficiency at large spatial scales \citep{read_process-guided_2019,ouguz2023survey,kratzert2019toward_ung,sun2021explore}. Specifically, deep learning architectures like long short-term memory (LSTM) networks \change{are}{have been} increasingly used for time series predictions due to their ability to model systems and variables that have memory, i.e. where past conditions influence present behavior  (e.g., snowpack depth; \citep{lees2022hydrological}). LSTMs have shown to outperform both state-of-the-art process-based models and also classical ML models (e.g., XGBoost, random forests, SVM) for applications like lake temperature \citep{read_process-guided_2019,jia_physics-guided_2021,daw2020physics}, stream temperature \citep{weierbach2022stream,feigl_machine-learning_2021}, and groundwater dynamics \citep{jing2022comparison} predictions amongst many others. Other deep learning architectures effective for time series modeling, but seen less often in hydrology, include the simpler gated recurrent unit (GRU) \citep{chung2014empirical} or more recent innovations like the temporal convolution network (TCN) \citep{lea2017temporal}, or spatiotemporally aware process-guided deep learning models \citep{topp2023stream}. \add{Recent advancements have also introduced transformer-based methods \protect\citep{yin2022rr}, which are architecturally able to model long-term dependencies more effectively than LSTM \protect\citep{zeyer2019comparison,wen2022transformers}. Transformers have been recently shown to occasionally outperform other methods for streamflow prediction \protect\citep{amanambu2022hydrological,yin2023runoff,xu2023deep}. However, so far, these alternatives to LSTM} have primarily focused on temporal predictions in well-monitored locations.

Understanding how to leverage state-of-the-art ML with existing observational data for \change{PUBs}{prediction in unmonitored sites} can lend insights into both model selection and training for transfer to new regions, as well as sampling design for new monitoring paradigms to optimally collect data for modeling and analysis. However, to date, ML-based approaches have not been sufficiently compared or benchmarked, making it challenging for researchers to determine which architecture to use for a given prediction task.  In this paper, we provide a comprehensive and systematic review of ML-based techniques for time series \change{PUBs}{predictions in unmonitored sites} and demonstrate their use for different environmental applications. We also enumerate the gaps and opportunities that exist for advancing research in this promising direction. The scope of our study is limited to using ML \change {to transfer predictions to new locations lacking any observations of the target environmental variable}{for predictions in unmonitored scenarios as defined above.} \remove{We do not cover temporal predictions at monitored locations with sparse data, i.e., where few data points are available, which is a different problem where innovative ML approaches such as few-shot learning are making significant progress (e.g. (Ghosh et al. 2022)), including in hydrological applications such as streamflow (Chen et al. 2022) and stream temperature (Jia et al. (2022)) predictions.} We \remove{also} do not cover the many statistical and ML-based efforts recent years for regionalizing process-based hydrological models, a topic that is covered extensively in the recent review  \citet{guo_regionalization_2020}.  \change{Lastly, w}{W}e also exclude remote sensing applications to \change{either estimate key  environmental variables or parameters for  model calibration. Although the use of remote sensing for monitoring at large spatial scales is a promising future direction for estimation of}{estimate} variables at previously unmonitored inland water bodies\change{, t}{. This is a different class of problems and t}here are significant challenges to increasing the scale and robustness of remote sensing applications including atmospheric effects, measurement frequency, and insufficient resolution for smaller water bodies like rivers \citep{topp2020research}, which are detailed in a number of reviews \citep{odermatt2012review,gholizadeh2016comprehensive,giardino2019imaging, topp2020research}. 

We organize the paper as follows. Section \ref{sect:frameworks} first describes different ML and knowledge-guided ML frameworks that have been applied for water resources time series \change{PUBs}{predictions in unmonitored sites}. Then, Section \ref{sect:discussion} summarizes and discusses overarching themes between methods, applications, regions, and datasets. Lastly, Section \ref{sect:open_questions} analyzes the gaps in knowledge and lists open questions for future research. 

\section{Machine Learning Frameworks for Predictions in Unmonitored Sites}
\label{sect:frameworks}
In this section\add{,} we describe different ML methodologies that have been used for \remove{PUB} applications in water resources time series modeling \add{for unmonitored sites}. Generally, the process of developing \change{PUB}{these} ML models first involves generating predictions for a set of entities (e.g. stream gauge sites, lakes) with monitoring data of the target variable (e.g. discharge, water quality). Then, the knowledge, data, or models developed on those systems are used to predict the target variable on entities with no monitoring data available. Importantly, for evaluation purposes these models are often tested on \textit{pseudo} unmonitored sites, where data is withheld until the testing stage to mimic model building for real unmonitored sites. 

The most commonly used type of model for this approach is known as an entity-aware model \citep{kratzert2019towards_og,ghosh2023entity}\footnote{In this paper we use the term "entity-aware" in the context of a general way of modeling a large number of entities with inherent characteristics with ML, as opposed to the "entity-aware long short-term memory" architecture in \citet{kratzert2019towards_og}.}, which attempts to incorporate inherent characteristics of different entities to improve prediction performance. These characteristics across the literature are also called attributes, traits, or properties. The concept is similar to trait-based modeling to map characteristics to function in ecology and other earth sciences \citep{zakharova2019trait}. The underlying assumption is that the input data used for prediction consists of both dynamic physical drivers (e.g. daily meteorology) and site-specific characteristics of each entity such as their geomorphology, climatology, land cover, or land use. Varied ML methodologies have been developed that differ both in how these characteristics are used to improve performance and also how entities are selected and used for modeling. These approaches are are described further below and include building a single model using all available entities or subgroups of entities deemed relevant to the target unmonitored sites (Section \ref{subsect:broad_scale}), transfer learning of models from well-monitored sites to target sites (Section \ref{subsec:transfer_learn}), and a cross-cutting theme of integrating ML with domain knowledge and process-based models (Section \ref{subsect:kgml}). 

\subsection{Broad-scale models using all available entities or a subgroup of entities}
\label{subsect:broad_scale}
Typically, process-based models have been applied and calibrated to specific locations, which is fundamentally different from the ML approach of building a single regionalized model on a large number of sites (hence referred to as a broad-scale model) that inherently differentiates between dynamic behaviors and characteristics of different sites \citep{guo2021regionalization,golian2021regionalization}. The objective of broad-scale modeling is to learn and encode these differences such that differences in site characteristics translate into appropriately heterogeneous hydrologic behavior. Usually the choice is made to include \textit{all} possible sites or entities in building a single broad-scale model. However, using the entirety of available data is not always optimal. Researchers may also consider selecting only a subset of entities for training for a variety of reasons including (1) the entire dataset may be imbalanced such that performance diminishes on minority system types \citep{wilson2020achieving}, (2) some types of entities may be noisy, contain erroneous or outlier data, or have varying amount of input data, or (3) to save on the computational expense of  building a broad-scale model. Traditionally in geoscientific disciplines like hydrology, stratifying a large domain of entities into multiple homogeneous subgroups or regions that are "similar" is common practice. This is based on evidence in process-based modeling that grouping heterogeneous sites for regionalization can negatively affect performance when extrapolating to unmonitored sites \citep{lettenmaier1987effect,hosking1997regional}.  Therefore, it remains an open question whether using all the available data is the optimal approach for building training datasets for \change{PUBs}{predictions in unmonitored sites}. Copious research has been done investigating various homogeneity criteria  trying to find the best way to group sites for these regionalization attempts for process-based modeling \citep{burn1990evaluation,burn1990appraisal,guo2021regionalization}, and many recent approaches also leverage ML for clustering sites (e.g. using k-means \citep{tongal2017cross,aytacc2020unsupervised}) prior to parameter regionalization \citep{sharghi2018application,guo2021regionalization}. 

Many studies use subgroups of sites when building broad-scale models using ML. For example, \citet{araza2020data} demonstrate that a principal components analysis-based clustering of 21 watersheds in the Luzon region of the Philippines outperforms an entity-aware broad-scale model built on all sites together for daily streamflow prediction. Futhermore, \citet{weierbach2022stream} find that an ML model combining two regions of data in the United States for stream temperature prediction didn't perform better than building models for each individual region. \citet{chen2020estimating} cluster weather stations by mean climatic characteristics when building LSTM and temporal convolution network models for predicting evapotranspiration in out-of-sample sites, claiming models performed better on similar climatic conditions. Additionally for stream water level prediction in unmonitored sites, \citet{corns2022deep} group sites based on the distance to upstream and downstream gauges to include proximity to a monitoring station as a criteria for input data selection. The water levels from the upstream and downstream gauges are also used as input variables. The peak flood prediction model described in Section \ref{subsect:broad_scale} divides the models and data across the 18 hydrological regions in the conterminous US as defined by USGS \citep{us_geological_survey_usgs_1994}. 

However, it remains to be seen how selecting a subgroup of entities as opposed to using all available data fairs in different prediction applications because much of this work does not compare the performances of both these cases. When viewed through the lens of modern data-driven modeling, evidence suggests deep learning methods in particular may benefit from pooling large amounts of heterogeneous training data. \citet{fang2022data} demonstrate this effect of "data synergy" on both streamflow and soil moisture modeling in gauged basins showing that deep learning models perform better when fed a diverse training dataset spanning multiple regions as opposed to homogeneous dataset on a single region even when the homogeneous data is more relevant to the testing dataset and the training datasets are the same size. \add{A recent opinion piece \protect\citet{kratzert2024hess} also make an argument against building deep learning models, specifically LSTM models, on streamflow data from small homogeneous sets of watersheds, especially for predicting in umonitored areas and for extreme events.} Moreover in \citet{willard2023machine}, regional LSTM models of stream temperature in the United States performed worse than the LSTM model built on all sites in the CONUS for 15 out of 17 regions, and single-site trained models transferred to the testing sites generally performed worse except when pre-trained on the global model. 

Overall across broad-scale modeling efforts, studies differ in how the ML framework leverages the site characteristics. The following subsections describe different approaches of incorporating site characteristics into broad-scale models that use all available entities or a subgroup, covering direct concatenation of site characteristics and dynamic features, encoding of characteristics using ML, and the use of graph neural networks to encode dependencies between sites.

\subsubsection{Direct concatenation broad-scale model} 
\label{subsubsec:direct_concat}
\begin{figure}[t]
\FIG{\includegraphics[width=.48\textwidth]{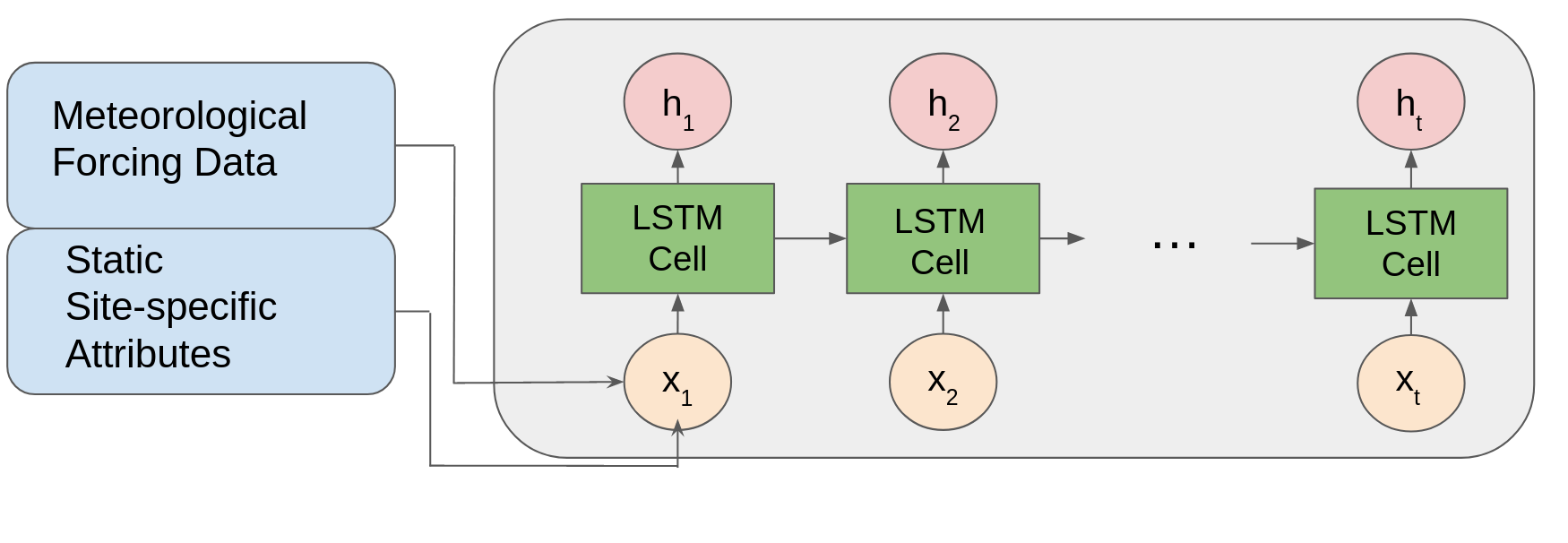}}
{\caption{Example of an long short-term memory (LSTM) network model with directly concatenated site characteristics and dynamic inputs}
\label{fig:ealstm}}
\end{figure}

When aggregating data across many sites for a entity-aware broad-scale model, it is common to append site characteristics directly with the input forcing data directly before feeding it to the ML model. Shown visually in Figure \ref{fig:ealstm}, this is a simple approach that does not require novel ML architecture, and is therefore very accessible for researchers. Although some characteristics can change over time, many applications treat these characteristics as static values over each timestep through this concatenation process, even though commonly used recurrent neural network-based approaches like LSTM are not built to incorporate static inputs \citep{lin2018early,rahman2020predicting,li2021integrating}. In a landmark result for temporal streamflow predictions, \citet{kratzert2019towards_og} used an LSTM with directly concatenated site characteristics and dynamic inputs built on 531 geographically diverse catchments within the Catchment Attributes and Meteorology for Large-sample Studies (CAMELS) dataset, and were able to predict more accurately on unseen data on the same 531 test sites than state-of-the-art process-based models calibrated to each basin individually. Given the success of the model, that study was expanded to the scenario of predicting in unmonitored stream sites \citep{kratzert2019toward_ung}, where they found the accuracy of the broad-scale LSTM with concatenated features in ungauged basins was comparable to calibrated process-based models in gauged basins. \citet{arsenault2023continuous} and \citet{jiang_improving_2020} further show a similar broad-scale LSTM can outperform the state-of-the-art regionalization of process-based models for \change{PUBs}{predictions in ungauged basins} in the United States, and similar results are seen in Russian \citep{ayzel2020streamflow}, Brazilian \citep{nogueira2022deep}, and Korean \citep{choi2022utilization} watersheds. \add{More recently, attention-based transformer models have been used in \protect\citet{yin2023runoff} for streamflow prediction on the CAMELS dataset showing improved performance over multiple kinds of LSTM models for both prediction in individual ungauged sites and entire ungauged regions.} Broad scale models have also  used for prediction of other environmental variables like continental-scale snow pack dynamics \citep{wang2022exploring}, monthly baseflow \citep{xie2022estimating}, dissolved oxygen in streams \citep{zhi2021hydrometeorology}, and lake surface temperature \citep{willard_daily_2022}.

The previously mentioned approaches in most cases focus on predicting mean daily values, but accurate predictions of extremes (e.g. very high flow events or droughts) remains an outstanding and challenging problem in complex spatiotemporal systems \citep{jiang2022predicting}. This is a longstanding fundamental challenge in catchment hydrology \citep{salinas2013comparative}, where typically the approach has been to subdivide the study area into fixed, contiguous regions that are used to regionalize predictions for floods or low flows from process-based models for all catchments in a given area. At least for process-based models, this has been shown to be more successful than global regionalizations \citep{salinas2013comparative}. As recent ML and statistical methods are shown to outperform process-based models for the prediction of extremes \citep{frame2022deep,viglione2013comparative}, opportunities exist to apply broad-scale entity-aware methods in the same way as daily averaged predictions. Challenges facing ML models for extremes prediction include replacing common loss functions like mean squared error which tend to prioritize average behavior and may not adequately capture the rare and extreme events \citep{mudigonda2021deep}, and dealing with the common scenario of extremes data being sparse \citep{zhang2011indices}.  Initial studies using broad-scale models with concatenated inputs for peak flood prediction show that these methods can also be used to predict extremes. For instance, \citet{rasheed2022advancing} built a peak flow prediction model that combines a "detector" LSTM that determines if the meteorological conditions pose a flood risk, with an entity-aware ML model for peak flow prediction to be applied if there is a risk. They show that building a model only on peak flows and combining it with a detector model improves performance over the broad-scale LSTM model trained to predict mean daily flows (e.g. \citet{kratzert2019toward_ung}). Though initial studies like this show promise, further research is required to compare techniques that deal with the imbalanced data, i.e. extreme events are often rare outliers, different loss function and evaluation metrics for extremes, and different ML architectures.  


Based on these results, it appears as though site characteristics can contain sufficient information to differentiate between site-specific dynamic behaviors for a variety of prediction tasks. This challenges a longstanding hydrological perspective that transferring models and knowledge from one basin to another requires that they must be functionally similar \citep{razavi2013streamflow,guo2021regionalization,fang2022data}, since these broad-scale models are built on a large number of heterogeneous sites. A recent study \citet{li2022regionalization} also substitutes random values as a substitute for site characteristics in a direct concatenation broad-scale LSTM to improve performance and promote entity-awareness in the case of missing or uncertain characteristics. 

\subsubsection{Concatenation of encoded site characteristics for broad-scale models}
\label{subsubsec:encoding}
\begin{figure}[H]
\caption{Example of a combination static feature encoder neural network with a long short-term memory (LSTM) network model}
\centering
\includegraphics[width=.78\textwidth]{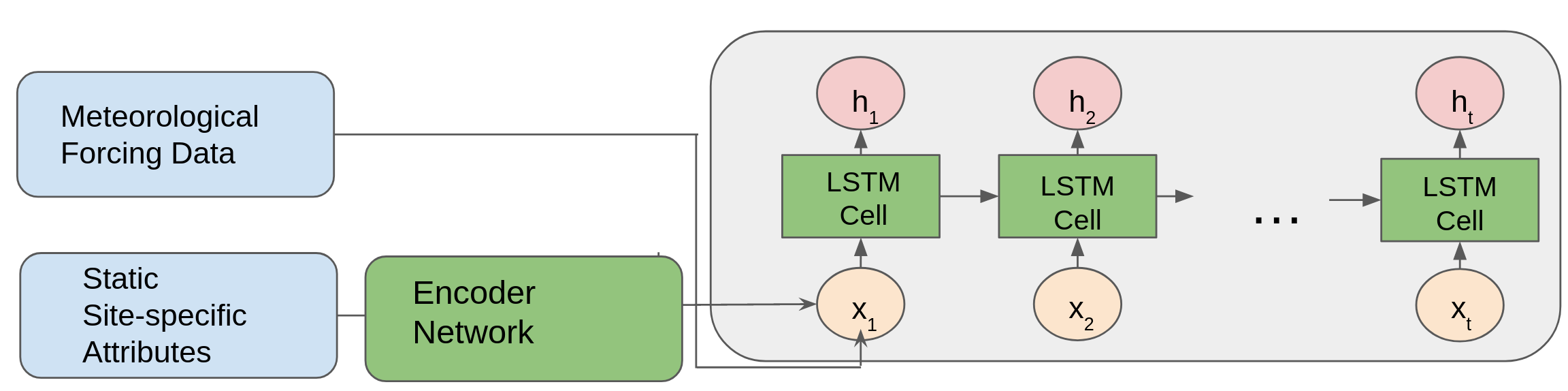}
\label{fig:encoder_lstm}
\end{figure}

Though recurrent neural network models like the LSTM have been used with direct concatenation of static and dynamic features, other methods have been developed that encode watershed characteristics as static features to improve accuracy or increase efficiency. Shown in Figure \ref{fig:encoder_lstm}, one approach is to use two separate neural networks, where the first learns a representation of the "static" characteristics using an encoding neural network (e.g. an autoencoder), and the second takes that encoded representation at each time-step along with dynamic time-series inputs to predict the target using a time series ML framework (e.g. LSTM). \add{This has been shown to be effective mostly in healthcare data applications \protect\citep{esteban2016predicting,li2021integrating,lin2018early}, but also in lake temperature prediction in \citet{tayal2022invertibility}}. The idea is to extract the information from characteristics that accounts for data heterogeneity across multiple entities. This extraction process is independent from the LSTM or similar time series model handing the dynamic input, and therefore can be flexible in how the two components are connected. Examples to improve efficiency include, (1) static information may not be needed at every time step and be applied only at the time step of interest \citep{lin2018early}, or (2) the encoding network can be used to reduce the dimension of static features prior to connecting with the ML framework doing the dynamic prediction \citep{kao2021fusing}. In terms of performance, works from multiple disciplines have found these type of approaches improve accuracy over the previously described direct concatenation approach \citep{lin2018early,tayal2022invertibility,rahman2020predicting}. 

\change{Although primarily demonstrated for healthcare applications, the use of an additional encoding network has been seen in hydrological studies.}{In water resources applications, } \citet{tayal2022invertibility} demonstrate this in lake temperature prediction using an invertible neural network in the encoding step, showing slight improvement over the static and dynamic concatenation approach. Invertible neural networks have the ability to model forward and backwards processes within a single network in order to solve inverse problems. For example, their model uses lake characteristics and meteorological data to predict lake temperature, but can also attempt to derive lake characteristics from lake temperature data. It has also been shown in streamflow prediction that this type of encoder network can be used either on the site characteristics \citep{jiang_improving_2020} or also on partially available soft data like soil moisture or flow duration curves \citep{feng2021mitigating}. In \citet{jiang_improving_2020}, they include a feed-forward neural network to process static catchment-specific attributes separately from dynamic meteorological data prior to predicting with a physics-informed neural network model. However, it is not directly compared with a model using the static features without any processing in a separate neural network so the added benefit is unclear. \citet{feng2021mitigating} further show an encoder network to encode soil moisture data if it is available prior to predicting streamflow with an LSTM model, but show limited benefit over not including the soil moisture data. 

\subsubsection{Broad-scale graph neural networks}
\label{subsubsect:spatial_context}

The majority of works in this study treat entities as systems that exist independently from each other (e.g. different lakes, different stream networks). However, many environmental and geospatial modeling applications exhibit strong dependencies and coherence between systems \citep{reichstein_deep_2019}. These dependencies can be real, interactive physical connections, or a coherence in dynamics due to certain similarities regardless of whether the entities interact. For example, water temperature in streams is affected by a combination of natural and human-involved processes including meteorology, interactions between connected stream segments within stream networks, and the process of water management and timed release reservoirs. Similar watersheds, basins, or lakes may also exhibit dependencies and coherence based on characteristics or climatic factors \citep{george2000factors,magnuson1990temporal,huntington2003historical,kingston2006river}. Popular methods like the previously described broad-scale models using direct concatenation of inputs (Section \ref{subsubsec:direct_concat}) offer no intuitive way to encode interdependencies between entities (e.g. in connected stream network) and often ignore these effects. Researchers are beginning to explore different ways to encode these dependencies explicitly by using graph neural networks (GNNs) for broad-scale modeling of many entities. The use of GNNs can allow the modeling of complex relationships and interdependencies between entities, something traditional feed-forward or recurrent neural networks cannot do \citep{wu2020comprehensive}.  GNNs have seen a surge in popularity in recent years for many scientific applications and several extensive surveys of GNNs are available in the literature \citep{bronstein2017geometric,battaglia2018relational,zhou2020graph,wu2020comprehensive}. Hydrological processes naturally have both spatial and temporal components, and GNNs attempt to exploit the spatial connections, causative relations, or dependencies between similar entities analogous to the way that the LSTM architecture exploits temporal patterns and dependencies. Recent work has attempted to encode stream network structure within GNNs to capture spatial and hydrological dependencies for applications like drainage pattern recognition \citep{yu2022recognition}, groundwater level prediction \citep{bai2022graph}, rainfall-runoff or streamflow prediction \citep{zhao2020joint,kratzert2021large,sit2021short,sun2021explore,feng2022graph,kazadi2022flood}, lake temperature prediction \citep{stalder2021probabilistic}, and stream temperature prediction \citep{chen2021heterogeneous,bao_partial_2021,chen2022physics}. 


In hydrology, there are \change{two}{three} intuitive methods for construction of the graph itself. The first is geared towards non-interacting entities, building the graph in the form of pair-wise similarity between entities, whether that be between site characteristics \citep{sun2021explore}, spatial locations \citep{zhang2021sea,sun2021time} (e.g. latitude/longitude), or both \citep{xiang2021high}. The second type is geared more toward physically interacting entities, for example the upstream and downstream connections between different stream segments in a river network \citep{jia_physics-guided_2021} or connections between reservoirs with timed water releases to downstream segments \citep{chen_heterogeneous_2021}. \add{The third type starts with an \textit{a priori} connectivity matrix like the previous type, but lets the GNN learn an adaptive connectivity matrix during training based on the sites' dynamic inputs, attributes, or location \protect\citep{sun2022graph}.} Relying solely on the characteristics or location for graph construction in the non-interacting case more easily allows for broad-scale modeling because it can model spatially disconnected entities, however it introduces no \textit{new} information (e.g. physical connectivity) beyond what the previously described direct concatenation-based methods use since the static characteristics would be the same. However, performance could still improve and interpretations of encodings within a graph framework could yield new scientific discovery since pairwise encodings between entities can be directly extracted. Graphs built using real physical connections between entities (e.g. stream segments in a stream graph), on the other hand, allow for the capability to learn how information is routed through the graph and how different entities physically interact with each other. So far, this has only been seen on stream modeling using stream network graphs \citep{jia_physics-guided_2021,kratzert2021large,bindas2020routing,topp2023stream}. \add{The third type is useful when combining the physical connectivity between sites with similarity in inputs, and also in cases where the inputs are at a different scale than the target variable, e.g. when meteorological variables are at kilometer scale and streamflow is at point scale.}

\begin{figure}[h]
\caption{Conceptual example of transductive and inductive graph learning. In both left and right panels, $\mathscr{F}$ is a model learned during training. Blue and red nodes represent entities with data for use in training and test entities without any data respectively. In transductive graph learning, the model has access to nodes and edges associated with test entities during training, but no new nodes can be introduced during testing. In inductive graph learning, the model is trained on an initial graph without any knowledge of the test entities, but the model can generalize to any new nodes during testing. 
 }
\centering
\includegraphics[width=\textwidth]{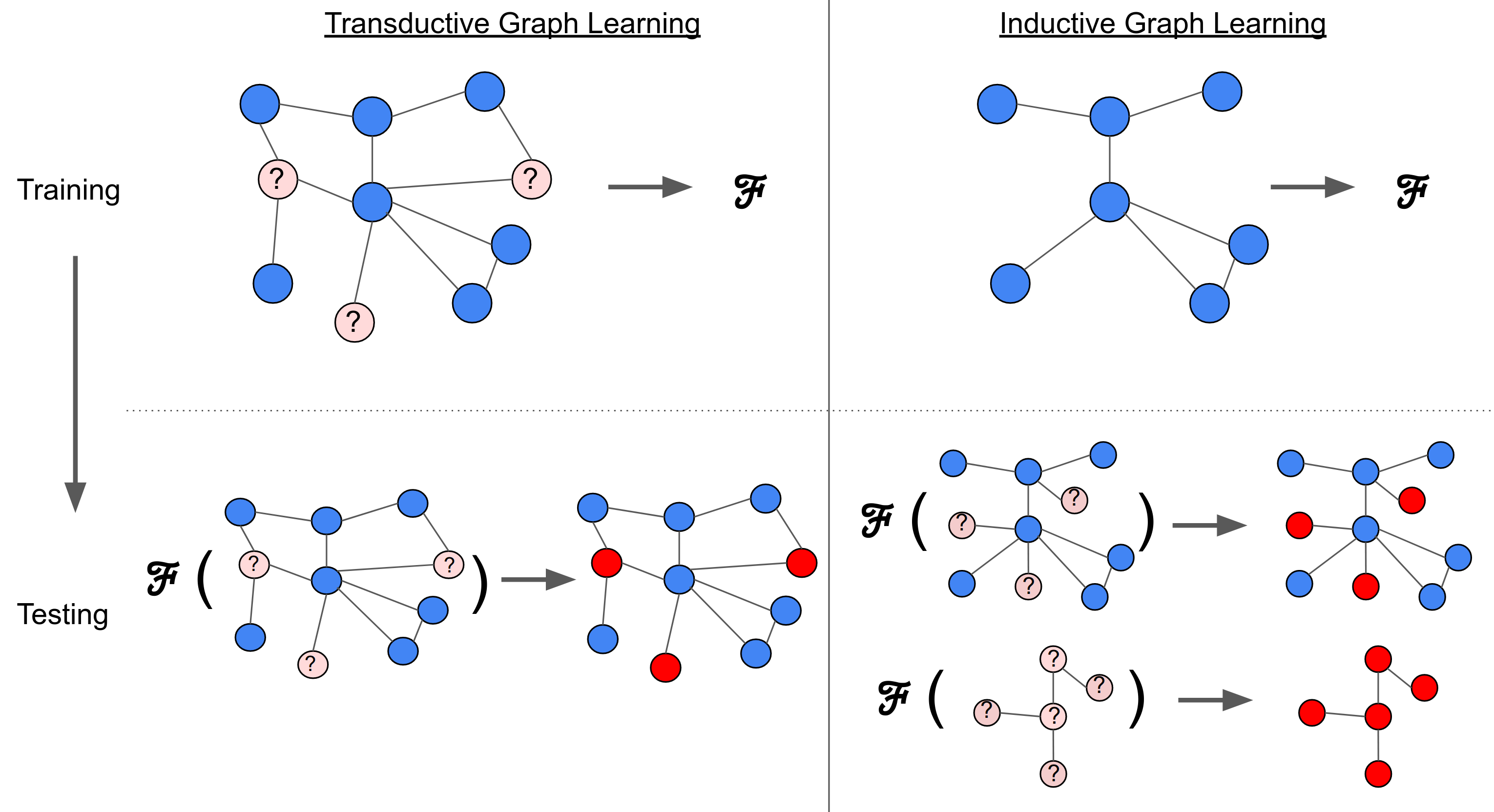}
\label{fig:graph_concept}
\end{figure}

 There are two different classes of GNN models, transductive and inductive, which differ in how the graph is incorporated in the learning process. Depending on how the graphs are constructed, one of these is more natural than the other. A conceptual depiction of both is shown in Figure \ref{fig:graph_concept}. The key aspect of transductive GNNs is that both training and testing entities must be present in the graph during training. A prerequisite for this approach is that the test data (e.g. input features in unmonitored sites) is available during model training, and one key aspect is that the model would need to be completely re-trained upon the introduction of new test data. Even if the training data is unchanged prior to re-training, introducing new test nodes in the graph  can affect how information is diffused to each training node during optimization \citep{ciano2021inductive}.  This type of approach is generally preferred for river network modeling given the often unchanging spatial topology of the sub-basin structure which is known \textit{a priori} \citep{jia_physics-guided_2021,sit2021short,moshe2020hydronets}. Graph connections from the test nodes to the training nodes in a transductive setting can be used either in the training or prediction phase, or both \citep{rossi2018inductive}. Inductive GNNs on the other hand, are built using only training entities and allow for new entity nodes to be integrated during testing. For applications that continuously need to predict on new test data, inductive approaches are much more preferred. New entity nodes are able to be incorporated because inductive frameworks also learn an information aggregator that transfers the necessary information from similar or nearby nodes to predict at nodes unseen during training. However, this also means connections between nodes only present in the test data and those in the training data are unseen during model training as opposed transductive approaches where they are included. As shown in Figure \ref{fig:graph_concept}, inductive graph learning can either be done on nodes that connect with training set nodes in the graph or those that are disconnected. Inductive GNNs can be understood as in the same class as more standard supervised ML models like LSTM or feed-forward neural networks, where they are able to continuously predict on new test data without the need for re-training. 

A few studies use GNNs for \change{PUB predictions in }{prediction in unmonitored sites for} water resources applications. \citet{sun2021explore} use different types of spatiotemporal GNNs including three transductive GNN methods, two variants of the ChebNet-LSTM \citep{yan2021spatial} and a Graph Convolutional Network LSTM (GCN-LSTM) \citep{seo2018structured}, compared with a GNN that can used as either transductive or inductive, GraphWaveNet \citep{wu2019graph}.  In all cases, the graph is initially  constructed as an adjacency matrix containing the pairwise Euclidean distance between stream sites using site characteristics. Importantly, all four models simplify to direct concatenation-based models described in Section \ref{subsect:broad_scale} if the graph convolution-based components are removed (See Figure S2 in \citet{sun2021explore} for a visualization). For ChebNet-LSTM and GCN-LSTM, the direct concatenation-approach would effectively simplify the architecture to a traditional LSTM, and for GraphWaveNet, it would simplify to a gated temporal convolution network (TCN). They found that the for the transductive case, both ChebNet-LSTMs and GCN-LSTM performed worse in terms of median performance across basins than the standard LSTM and GraphWaveNet was the only one that performed better. GraphWaveNet, the only GNN also capable of doing inductive learning, also performed better in the inductive case than standard LSTM.  \citet{jia_physics-guided_2021} take a different spatiotemporal GNN approach for stream temperature temporal predictions, where they construct their graph by using stream reach lengths with upstream and downstream connections to construct a weighted adjacency matrix. They found their GNN pre-trained on simulation data from the PRMS-SNTemp process-based model \citep{markstrom2012p2s} outperformed both a non-pre-trained GNN and a baseline LSTM model. Based on these results, we see that encoding dependencies based on site characteristics as well as physical interaction and stream connections within GNNs, can improve performance over existing deep learning models like the feed-forward artificial neural network (ANN) or LSTM. 

Some studies have explored different ways of constructing the adjacency matrix based on the application and available data. An example of a domain-informed method for graph construction can be seen in \citet{bao_partial_2021} for stream temperature \change{PUBs}{predictions in unmonitored sites}, where they leverage partial differential equations of underlying heat transfer processes to estimate the graph structure dynamically. This graph structure is combined with temporal recurrent layers to improve prediction performance beyond existing process-based and ML approaches. Dynamic temporal graph structures like this are common in other disciplines like social media analysis and recommender systems, but have not been widely used in the geosciences \citep{longa2023graph}.

\subsection{Transfer learning}
\label{subsec:transfer_learn}
Transfer learning is a powerful technique for applying knowledge learned from one problem domain to another, typically to compensate for missing, nonexistent, or unrepresentative data in the new problem domain. The idea is to transfer knowledge from an auxiliary task, i.e., the source system, where adequate data is available, to a new but related task, i.e., the target system, often where data is scarce or absent \citep{pan2009survey, weiss2016survey}. Situations where transfer learning may be more desirable than broad-scale modeling include when (1) a set of highly tuned and reliable source models (ML, process-based or hybrid) may already be available, (2) local source models are more feasible computationally or more accurate than broad-scale models when applied to unmonitored systems, or (3) broad-scale models may need to be transferred and fine tuned to a given region or system type more similar to an unmonitored system. In the context of geoscientific modeling, transfer learning for ML is analogous to calibrating process-based models in well-monitored systems and transferring the calibrated parameters to models for unmonitored systems, which has shown success in hydrological applications \citep{kumar2013implications,roth2016model}. Deep learning is particularly amenable to transfer learning because it can make use of massive datasets from related problems and alleviate data paucity issues common in applying data-hungry deep neural networks to environmental applications \citep{shen_transdisciplinary_2018,naeini2019transfer}. Transfer learning using deep learning has shown recent success in water applications such as flood prediction \citep{kimura2019convolutional,zhao2021improving}, soil moisture \citep{li2021improved}, and lake and estuary water quality \citep{tian2019transfer,willard_predicting_2021}. 

\begin{figure}[t]
\caption{Process diagram of the Meta Transfer Learning framework. Models are first built from data-rich source domains. The metamodel is trained using characteristics extracted from the source domains to predict the performance metrics from transferring models between source domains. Then given a target system or domain, the metamodel is able to output a prediction of how well each of the source models will perform on the target system. Adapted from \protect\citet{willard_predicting_2021}}.
\centering
\includegraphics[width=.45\textwidth]{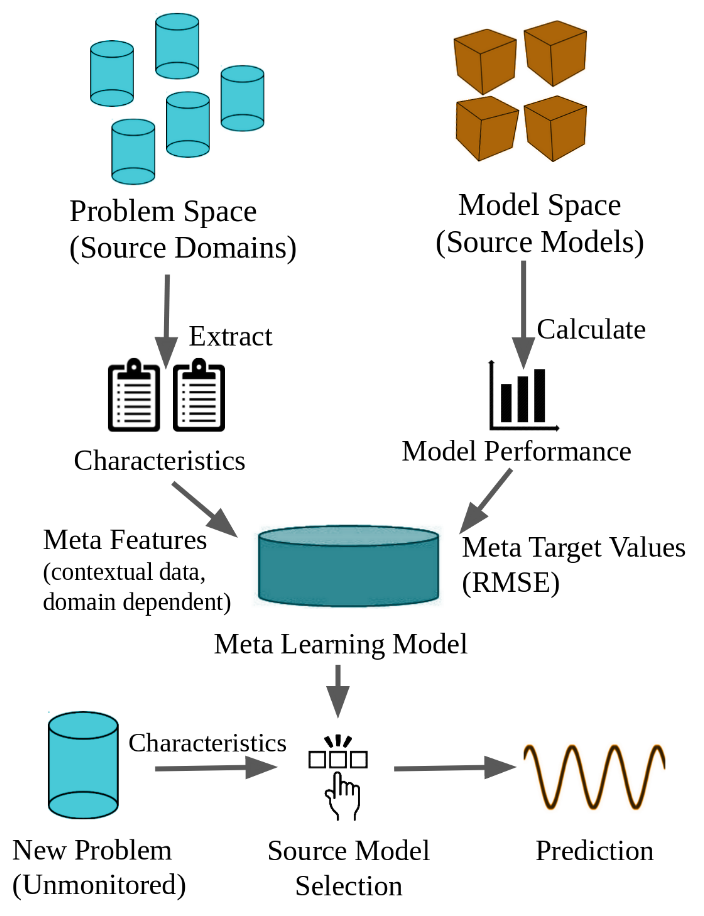}
\label{fig:mtl}
\end{figure}

Transfer learning can also be a capable tool for \change{PUBs}{predictions in unmonitored sites} \citep{tabas2021structure}, although most applications  typically assume that some data is available in the target system for fine-tuning a model , which is often referred to as few shot learning with sparse data \citep{weiss2016survey,zhuang2020comprehensive}. The specific case of transferring to a system or task without any training data is also known as "zero-shot learning"  \citep{romera2015embarrassingly}, where only the inputs or a high level description may be available for the testing domain that does not contain any target variable values. This is a significantly more challenging problem because taking a pre-trained model from a data-rich source system and fine-tuning it on the target system is not possible, and instead other contextual data about the source and target systems must be used. For the case of \change{PUBs}{unmonitored prediction}, we often only have the dynamic forcing data and the characteristics of the target system (site) available. The following subsections cover different ways researchers have addressed the zero-shot transfer learning problem for water resources prediction.

\subsubsection{Choosing which model to transfer}
\label{subsubsec:tl_model_choice}
A central challenge in zero-shot transfer learning is determining which model to transfer from a related known task or how to build a transferable model. Previous work on streamflow prediction has based this purely on expert knowledge. For example, \citet{singh2022streamflow} operates under the assumption that the model must be trained on other basins in the same climatic zone and at least some of the source basin's geographical area must have similar meteorological conditions to the target basin. Other work has transferred models from data-rich regions to data-poor regions without any analysis of the similarity between the source and target regions. For example, \citet{le2022streamflow} transfer ML streamflow models built on North America (987 catchments), South America (813 catchments), and Western Europe (457 catchments); to data-poor South Africa and Central Asian regions. They transfer these models as-is and do not take into account any of the sparse data in the data-poor region or the similarity between regions and find that the local models trained on minimal data outperform the models from data-rich regions. Attempts have also been made to use simple expert-created distance-based metrics (e.g. \citet{burn1993estimation}) using the site characteristic values \citep{vaheddoost2023rainfall}. However, it is reasonable to think that a data-driven way to inform model building based off both the entity's characteristics and past modeling experiences would be possible. 

The idea of building or selecting a model by leveraging preexisting models is a type of \textit{meta-learning} \citep{lemke_metalearning:_2015,brazdil_metalearning:_2009}. More broadly meta-learning is the concept of algorithms learning from other algorithms, often in the task of selecting a model or learning how to best combine predictions from different models in the context of ensemble learning.  One meta-learning strategy for model selection is to build a metamodel to learn from both the model parameters of known tasks (with ground truth observations) and the correlation of known tasks to zero-shot tasks \citep{pal2019zero}. For example, in lake temperature modeling, \citet{willard_predicting_2021} use meta-learning for a model selection framework where a metamodel learns to predict the error of transferring a model built on a data-rich source lake to an unmonitored target lake. A diagram of the approach is shown in Figure \ref{fig:mtl}. They use variety of contextual data is used to make this prediction, including (1) characteristics of the lake (e.g. maximum depth, surface area, clarity etc., (2) meteorological statistics (e.g. average and standard deviation of air temperature, wind speed, humidity etc., (3) simulation statistics from an uncalibrated process-based model applied to both the source and target (e.g. differences in simulated lake stratification frequency), and (4) general observation statistics (e.g. number of training data points available on the source, average lake depth of measured temperature, etc). They show significantly improved performance predicting temperatures in 305 target lakes treated as unmonitored in the Upper Midwestern United States relative to the uncalibrated process-based General Lake Model \citep{hipsey_general_2019}, the previous state-of-the-art for broad-scale lake thermodynamic modeling. This was expanded to a streamflow application in \citet{ghosh2022meta} with numerous methodological adaptations. First, instead of using the site characteristics as is, they use a sequence autoencoder to create embeddings for all the stream locations by combining input time series data and simulated data generated by a process-based model. This adaptation alleviated a known issue in the dataset where the site characteristics were commonly incomplete and inaccurate. They also use a clustering loss function term in the sequence autoencoder to guide the model transfer, where source systems are selected based on available source systems within a given cluster of sites as opposed to building an ensemble with a set number of source sites. The clustering loss function term allows the model to learn a latent space that can correctly cluster river streams that can accurately transfer to one another. They show on streams within the Delaware River Basin that this outperforms the aforementioned simpler meta transfer learning frameworks on sites based on \citet{willard_predicting_2021}. In \citet{willard2023machine}, they expand on \citet{willard_predicting_2021} by also building a meta transfer learning framework that pre-trains each source model on CONUS-scale data, aiming to combine benefits of broad-scale modeling and site-specific transfer learning for the task of stream temperature prediction. They find a small performance improvement over the existing direct concatenation approach building a single model on all stream entities in the CONUS. 


\subsubsection{Fine-tuning models with sparse data}
\label{subsubsec:tl_localize_region}

\add{A common hydrologic prediction scenario is one in which broad-scale data and models are available but a target site has inadequate or sparse data. This is especially seen in remote, inaccessible, or under-monitored regions. Given a pre-trained model on broad-scale data or simulated process-based model outputs, fine-tuning ML models by adjusting parameters during a second training instance has the potential to improve the accuracy and relevance of the model for specific local conditions. Pre-training on process-based model outputs and fine-tuning on minimal sparse data has shown to be effective in lake temperature  \protect\citep{jia_physics-guided_2021-2,willard_predicting_2021} and stream temperature \protect\citep{jia_physics-guided_2021} prediction for as little as 0.1\% of available data to simulate a common prediction scenario where only a few measurements of the target variable may be available and show substantial increase in performance over an uncalibrated process-based model. Furthermore, in soil moisture prediction \protect\citet{li2021improved} show an effective pre-training on the large-scale process-based reanalysis ERA5-Land dataset \protect\citep{munoz-sabater_era5-land_2021} and fine-tuning on the smaller SMAP data \protect\citep{o2010nasa} showing increased explained variation of over 20\% compared to the non-fine-tuned version.}

Another transfer learning with fine-tuning strategy in geoscientific modeling that can also be based on pre-training is to localize a larger-scale or more data-rich regional or global model to a specific location or subregion. This variant of transfer learning has seen success in deep learning models for applications like soil spectroscopy \citep{padarian2019transfer,shen2022deep} and snow cover prediction \citep{guo2020extraction,wang2020estimating}. However, these strategies have seen mixed success in hydrological applications. \citet{wang2022exploring} show that localizing an LSTM predicting continental-scale snowpack dynamics to individual regions across the United States had insignificant benefit over the continental-scale LSTM. \citet{xiong2022predicting} show a similar result for the prediction of stream nitrogen export, where the individual models for the 7 distinct regions across the conterminous United States transferred to each other  did not outperform the continental-scale model using all the data. Also, \citet{lotsberg2021lstm} show that streamflow models trained on CAMELS-US (United States) transfer to CAMELS-GB (Great Britain) about as well as a model trained on the combined data from US and GB, and models trained on CAMELS-GB transfer to CAMELS-US about as well as a model using the combined data. They also show that the addition of site characteristics is not beneficial in transfer learning tasks, but acknowledge this could be due to the way data is normalized prior to training. Based on these results, it is possible that the entity-aware broad-scale model using all available data is already learning to differentiate between different regions or types of sites on its own and fine-tuning to more similar sites based on expert knowledge may be less useful. However, this remains to be demonstrated for most hydrological and water resources prediction tasks. \add{Other studies have also continued the practice of pre-training a model on a data-dense region like the US and fine-tuning on data-sparse regions like China \protect\citep{ma_transferring_2021,xu2023deep} or Kenya \protect\citep{oruche_transfer_2021}.}

\subsubsection{Unsupervised Domain Adaptation}
\label{subsubsec:tl_uda}
Domain adaptation methods are a subset of transfer learning algorithms that attempt to answer the question, \textit{how can a model both learn from a source domain and learn to generalize to a target domain}? Often domain adaptation seeks to minimize the risk of making errors on the target data, and not necessarily on the source data as in traditional supervised learning. Unsupervised domain adaptation (UDA), in particular, focuses on the zero-shot learning case of the target domain being void of target data. Similar to the types of graph neural networks mentioned in Section \ref{subsubsect:spatial_context}, review papers have divided transfer learning algorithms into the categories, (1) \textit{inductive} transfer learning where the source and target tasks are different and at least some labeled data from the target task is required to induce a model, (2) \textit{transductive} transfer learning where the source and target tasks are the same but from different feature space domains and zero labeled data is available from the target domain, and (3) \textit{unsupervised} transfer learning where no labeled data is available in both the source and target domains \citep{pan_survey_2010,niu2020decade}. UDA specifically lies in the transductive transfer learning scenario, and usually involves using the input data from the target or testing task during the training process, in addition to the source data. This aspect differentiates UDA from the previously described methods in this section. Researchers can employ different UDA methods when attempting to account for differences in the source and target tasks and datasets. Commonly UDA methods attempt to account for the difference in input feature distribution shifts between the source and task, but other methods attempt to account for the difference in distributions of labeled data.  This differs from previous approaches we have mentioned like the broad-scale models that generally ignore input data from testing sites, meta transfer learning that uses test data inputs during model selection but not during training, and localizing regional models which uses available data from regions containing the test sites but not any data from the test sites themselves. UDA has seen success in many disciplines including computer vision \citep{patel2015visual,csurka2017comprehensive}, robotics \citep{hoffman2016fcns,bousmalis2018using}, natural language processing \citep{blitzer2007biographies}, and fault diagnostics \citep{shi2022deep} but applications of UDA in hydrology are limited. In the only current hydrological example, \citet{zhou2022flooddan} introduce a UDA framework for unmonitored flood forecasting that involves a two-stage adversarial learning approach. The model is first pre-trained on a large sample source dataset, then they perform adversarial domain adaptation using an encoder to map the source and target inputs to the same feature space and learn the difference between the source and target datasets. They show this method is effective in flood forecasting across the Tunxi and Changhua flood datasets spanning Eastern China and Taiwan. Currently UDA that accounts for a shift in label distribution (real or synthetic) has not been attempted in hydrological prediction, and future research on UDA in hydrology will need to consider whether to account for either input or label distribution shift between entities and systems.



\subsection{Cross cutting theme: knowledge-guided machine learning}
\label{subsect:kgml}
There is a growing consensus that solutions to complex nonlinear environmental and engineering problems will require novel methodologies that are able to integrate traditional process-based modeling approaches with state-of-the-art ML techniques, known as \textit{Knowledge-guided machine learning} (KGML) \citep{karpatne2022knowledge} (also known as \textit{Physics-guided machine learning} or \textit{Physics-informed machine learning} \citep{willard_integrating_2022,muther2022physical,karpatne_theory-guided_2017}). These techniques have been demonstrated to improve prediction in many applications including lake temperature \citep{jia_physics-guided_2021-2,read_process-guided_2019}, streamflow \citep{herath2021hydrologically,hoedt_mc-lstm_2021,bhasme2022enhancing}, groundwater contamination \citep{soriano2021assessment}, and water cycle dynamics \citep{ng2021physics} among others. \citet{willard_integrating_2022} divide KGML methodologies into four classes; (i) physics-guided\footnote{In this paper, we use the term "knowledge-guided" as opposed to "physics-guided" but they are used interchangeably in the literature.} loss function, (ii) physics-guided initialization, (iii) physics-guided design of architecture, and (iv) hybrid physics-ML modeling. Many of these methods are helpful \change{in the case of PUBs}{for prediction in unmonitored sites}, since known physics or existing models can exist in the absence of observed target data. Note that KGML is a cross-cutting theme, as its principles can be integrated into either of the previously described broad-scale modeling and transfer learning approaches. The benefits we see from KGML as a class of standalone techniques can also help address resource efficiency issues in building both broad-scale entire-aware models and also source models in transfer learning while maintaining high predictive performance, training data efficiency, and interpretability relative to traditional ML approaches \citep{willard_integrating_2022}.

The field of KGML is rapidly advancing, and given the numerous applications we see for its use in hydrology, we include the following discussion on the different ways of harnessing KGML techniques in a given physical problem that has traditionally been simulated using process-based models. The following three subsections are divided based on how KGML techniques are used to either replace, augment, or recreate an existing process-based model. Section \ref{future_kgml} further expands on this discussion by addressing the role of KGML in the future of unmonitored prediction and open questions that exist.  

\subsubsection{Guiding ML with domain knowledge: KGML loss functions, architecture, and initialization}
Traditional process-based models for simulating environmental variables in complex systems do not capture all the processes involved which can lead to incomplete model structure (e.g., from simplified or missing physics). Though a key benefit of pure ML is the flexibility to literally fit any dataset as well as not being beholden to the causal structure that process-based models are, its inability to make use of process-based knowledge can lead to negative effects like sample inefficiency, inability to generalize to out-of-sample scenarios, and physically inconsistent solutions. When building an ML model as a replacement for a process-based model, there are at least three considerations to guide the ML model with domain knowledge for improved predictive performance; KGML loss function terms, architecture, and initialization. 

KGML loss function terms can constrain model outputs such that they conform to existing physical laws or governing equations. \add{In dynamical systems modeling and solving partial differentiable equations this technique is known as \protect\textit{physics-informed neural networks} (PINNs) pioneered by \protect\citet{raissi_physics-informed_2019}}. Steering ML predictions towards physically consistent outputs has numerous benefits.  For \change{PUBs}{prediction in unmonitored or data-sparse scenarios}, the major benefit of informed loss function terms is that often the computation requires little to no observation data. Therefore, optimizing for that term allows for the inclusion of unlabeled data in training, which is often the only data available. Other benefits include that the regularization by physical constraints can reduce the possible search space of parameters, and also potentially learning with fewer labeled data, while also ensuring the consistency with physical laws during optimization. KGML loss function terms have also shown that models following desired physical properties are more likely to be generalizable to out-of-sample scenarios \citep{read_process-guided_2019}, and thus become acceptable for use by domain scientists and stakeholders in water resources applications. Loss function terms corresponding to physical constraints are applicable across many different types of ML frameworks and objectives, however most of these applications have been in the monitored prediction scenario (e.g. lake temperature \citep{jia_physics-guided_2021-2,read_process-guided_2019,karpatne_physics-guided_2017}, lake phosphorous \citep{hanson_predicting_2020}, subsurface flow \citep{wang2020deep}). \add{We also see applications of PINNs in hydrology for solving PDEs for transmissivity \protect\citep{guo2023high}, solute transport \protect\citep{niu20231}, soil moisture \protect\citep{bandai2022forward}, groundwater flow \protect\citep{cuomo2023solving} and shallow water equations \protect\citep{nazari2022physics,feng2023physics}.} In this survey, we find only one work using informed loss function terms within a meta transfer learning framework for lake temperature modeling \citep{willard_predicting_2021} incorporating conservation of energy relating the ingoing and outgoing thermal fluxes into the lake. 

Another direction is to use domain knowledge to directly alter a neural network's architecture to implicitly encode physical consistency or other desired physical properties. However, KGML-driven architecture optimizing for physical consistency is usually understood as a hard constraint since the consistency is hardcoded into the model, whereas KGML loss function terms are a soft constraint that can depend on optimization and weights within the loss function.  Other benefits from KGML loss function terms are also experienced by KGML-driven model architecture, including reducing the search space and allowing for better out-of-sample generalizability. KGML-driven model  architectures have shown success in hydrology, however it has been limited to temporal predictions for monitored sites. Examples include \citet{jiang_improving_2020} where they show a rainfall-runoff process model can be embedded as a special recurrent neural layers in a deep learning architecture, \citet{daw_physics-aware_2019} where they show a physical intermediate neural network node as part of a monotonicity-preserving structure in the LSTM architecture for lake temperature, and more examples in the \citet{willard_integrating_2022} KGML survey. However, there is nothing preventing these approaches from being applied in the unmonitored scenario. 
 
Lastly, if process-based model output is already available, such as the National Water Model streamflow outputs \citep{noaa2016national}, FLake model lake surface temperature outputs within ERA5 \citep{munoz-sabater_era5-land_2021}, or PRMS-SNTemp simulated stream temperature \citep{markstrom2012p2s}, this data can be used to help pre-train an ML model, which is known as KGML initialization. In the \change{PUB}{unmonitored prediction} scenario, pre-training can be done on process-based model simulations of sites with no monitoring data. This is arguably the most accessible KGML method since there is no direct alteration of the ML approaches. By pre-training, the ML model can learn to emulate the process-based model prior to seeing training data in order to accelerate or improve the primary training. Numerous studies in water resources perform KGML-based model initialization by making use of process-based model output to inform ML model building, either to create site-specific embeddings used for similarity calculation in meta transfer learning  \citep{ghosh2022meta}, as a pre-training stage for source models in meta transfer learning \citep{willard_predicting_2021}, or as a pre-training stage for entity-aware broad-scale models \citep{koch2022long,noori2020water}.

\add{Beyond these traditional KGML approaches, there is also the concept of \protect\textit{neural operators}, which have emerged as a powerful class of ML capable of generalizing across different scenarios and scales. Unlike traditional neural networks that learn mappings between inputs and outputs with fixed dimensions, neural operators map between infinite-dimensional functional spaces \protect\citep{li2020fourier}.  While neural operators have not yet been directly applied to ungauged or unmonitored hydrologic time series prediction, recent studies demonstrate their potential in surrogate modeling of dynamical systems modeling for flood inundation \protect\citep{sun2023rapid}, geological carbon storage \protect\citep{tang2024multi}, and groundwater flow \protect\citep{taccari2023developing}. They also have the capability to increase computational efficiency within transformer architectures for scaling to high resolution or high dimensional data, specifically for vision transformers in  \protect\citet{guibas2021adaptive} and \citet{pathak2022fourcastnet}}.

\subsubsection{Augmenting process models with ML using hybrid process-ML models}
\label{subsubsect:hybrid}
In many cases certain aspects of process-based models may be sufficient but researchers seek to use ML in conjunction with an operating process-based to address key issues. Examples include where (1) process-based model outputs or intermediate variables are useful inputs to the ML model, (2) a process-based model may model certain intermediate variables better than others that could utilize the benefits of ML, or (3) optimal performance involves choosing between process-based models and ML models, based on prediction circumstances in real time. Using both the ML model and a process-based model \textit{simultaneously} is known as a hybrid process-ML model and is the most commonly used KGML technique for unmonitored prediction. In the \citet{willard_integrating_2022}  survey of KGML methods, they define hybrid models as either process and ML models working together for a prediction task, or a subcomponent of a process-based model being replaced by an ML model. This type of KGML method is also very accessible for domain scientists since it requires no alterations to existing ML frameworks. In this work, we do not cover the large body of work of ML predictions of process-based model parameters since these methods have been outpaced by ML for predictive performance and tend to extrapolate to new locations poorly \citep{nearing_what_2021}, but summaries can be found in \citet{reichstein_deep_2019} or  \citet{xu2021machine}.

The most common form of hybrid process-ML models in hydrological and water resources engineering is known as residual modeling. In residual modeling, a data-driven model is trained to predict a corrective term to the biased output of a process-based or mechanistic model.  This concept goes by other names such as error-correction modeling, model post-processing, error prediction, compensation prediction, and others. Correcting these residual errors and biases has been shown to improve the skill and reliability streamflow forecasting \citep{regonda2013short,cho2022improving}, water level prediction \citep{lopez2014alternative}, and groundwater prediction \citep{xu2015data}. When applying residual modeling to unmonitored prediction, the bias correcting ML model must be trained on either a large number of sites or sites similar to the target site. \citet{hales2022saber} demonstrate a framework to build a residual model for stream discharge prediction with the GEOGloWS ECMWF Streamflow Model that selects similar sites based on the dynamic time warping and euclidean distance time series similarity metrics. For unmonitored sites, they substitute simulated data instead of the observed data and show a substantial reduction in model bias in ungauged subbasins. 

A slight alteration to the residual model is a hybrid process-ML model that takes an ML model and adds the output of a process-based model as an additional input. This adds a degree of flexibility to the modeling process compared to the standard residual model as the residual error is not modeled explicitly, and multiple process-based model outputs can be used at once. \citet{karpatne_physics-guided_2017} showed that adding the simulated output of a process-based model as one input to an ML model along with input drivers used to drive the physics-based model for lake temperature modeling can improve predictions, and a similar result was seen in \citet{yang2019evaluation} augmenting flood simulation model based on prior global flood prediction models. This hybrid modeling approach has recently been applied to \change{PUBs}{unmonitored prediction} as well, with \citet{noori2020water} using the output of SWAT (Soil \& Water Assessment Tool \citep{arnold1998large}) as an input to a feed-forward neural network for predicting monthly nutrient load prediction in unmonitored watersheds. They find that the hybrid process-ML model has greater prediction skill in unmonitored sites than the SWAT model calibrated at each individual site. 

Another simple way to combine process-based models with ML models is through multi-model ensemble approaches that combines the predictions of two or more types models. Ensembles can both provide more robust prediction and allow quantification and reduction of uncertainty. Multiple studies in hydrology have shown that using two or more process-based models with different structures improves performance and reduce prediction uncertainty in ungauged basins \citep{cibin2014application,waseem2015ensemble}. \citet{razavi2016improving} show an ensemble of both ML models and process-based models for streamflow prediction, which further reduced prediction uncertainty and outperformed individual models. However this study is limited to building a model for an ungauged stream site using only the three most similar and closely located watersheds, as opposed to more comprehensive datasets like CAMELS. 

Comparisons between different types of hybrid models are not commonly seen, as most studies tend to use only one method. In one study highlighting different hybrid models, \citet{frame2021post} compare three approaches, (1) LSTM residual models correcting the National Water Model (NWM), (2) a hybrid process-ML model using an LSTM that takes the output of the NWM as an additional input, and (3) a broad-scale entity-aware LSTM like we have described in Section \ref{subsect:broad_scale}. They find that in the unmonitored scenario, the third approach performed the best, which leads to the conclusion that the output from the NWM actually impairs the model and prevents it from learning generalizable hydrological relationships. In many KGML applications, the underlying assumption is that the process-based model is capable of reasonably good predictions and adds value to the ML approaches.  Additional research is required to address when hybrid modeling is beneficial for unmonitored prediction, since there are often numerous process-based models and different ways to hybridize modeling for a given environmental variable.



\subsubsection{Building differentiable and learnable process-based models}
\label{subsubsec:dpb}
Numerous efforts have been made to build KGML models that have equal or greater accuracy than existing ML approaches but with increased interpretability, transparency, and explainability using the principles of differentiable process-based (DPB) modeling  \citep{khandelwal_physics_2020,feng2022differentiable,shen2023differentiable}. The main idea of DPB models is to keep an existing geoscientific model's structure but replace the entirety of its components with differentiable units (e.g. ML). From an ML point of view, it can be viewed as a domain-informed structural prior resulting in a modular neural network with physically meaningful components. This differs from the previously described hybrid process-ML methods that include non-differentiable process-based models or components. One recent example is shown in hydrological flow prediction in \citet{feng2022differentiable}, though similar models have been used in other applications like earth system models \citep{gelbrecht2022differentiable} and molecular dynamics \citep{alquraishi2021differentiable}. The DPB model proposed by \citet{feng2022differentiable} starts with a simple backbone hydrological model (Hydrologiska Byråns Vattenbalansavdelning model \citep{bergstrom1976development}), replaces parts of the model with neural networks, and couples it with a differentiable parameter learning framework (see Figure 1 in \citet{feng2022differentiable} for a visualization). Specifically, the process model structure is implemented as a custom neural network architecture that connects units in a way that encodes the key domain process descriptions, and an additional neural network is appended to the aforementioned process-based neural network model to learn the physical parameters. The key concept is that the entire framework is differentiable from end to end, and the authors further show that the model has nearly identical performance in gauged flow prediction to the record-holding entity-aware LSTM, while exhibiting interpretable physical processes and adherence to physical laws like conservation of mass. A simpler implementation is seen in \citet{khandelwal_physics_2020}, also for streamflow, where intermediate RNN models are used to predict important process model intermediate variables (e.g. snowpack, evapotranspiration) prior to the final output layer. In both of these implementations, we see a major advantage of the DPB model is the ability to output an entire suite of environmental variables in addition to the target streamflow variable, including baseflow, evapotranspiration, water storage, and soil moisture. The DPB approach has been further demonstrated on unmonitored prediction of hydrological flow in \citet{feng2022suitability}, showing better performance than the entity-aware LSTM for mean flow and high flow predictions but slightly worse for low flow. The results of DPB models in both unmonitored and monitored scenarios challenge the notion that process-based model structure rigidness is undesirable as opposed to the highly flexible nature of neural network, and that maybe elements of both can be beneficial when the performance is near-identical in these specific case studies. 
\begin{longtable}[h]
{|P{0.1\linewidth}  P{0.15\linewidth}  P{0.2\linewidth}  P{0.2\linewidth}  P{0.2\linewidth}|}

\captionsetup{width=\linewidth}
\caption{Literature Table. Abbreviations as follows, DCBS: direct concatenation broad-scale, TL: transfer learning ANN: artificial neural network (feed forward multilayer perceptron), GNN: graph neural network, LSTM: long short-term memory neural network, MARS: multi-adaptive regression splines, MLR: multilinear regression, GBR: gradient boosting regression, GRU: gated recurrent unit, PDE: partial differential equation, RF: random forest, SVR: support vector regression, TCN: temporal convolution network, XGB: extreme gradient boosting}\phantomsection\label{table:lit}
\\ \toprule

Work by & Variable Predicted & ML Framework Demonstrated & Type of Models Compared & Region Covered \\
\midrule
\endhead

\hline
\endfoot
\cite{araza2020data} 
    & Streamflow (daily) 
    & DCBS, subgroup of entities broad-scale model 
    & RF  
    & 21 watersheds in Luzon, Philippines 
 \\
\cite{arsenault2023continuous} 
    & Streamflow (daily) 
    & DCBS 
    & LSTM, 3 different process-based models (HSAMI, HMETS, GR4J) 
    & 148 catchments in Northeast North America 
 \\
\cite{ayzel2020streamflow} 
    & Streamflow (daily) 
    & DCBS 
    & LSTM, process-based models (GR4J) 
    & 200 catchments in Northwest Russia 
\\
\cite{bao_partial_2021} 
    & Streamflow (daily) 
    & KGML (PDE-driven graph network) 
    & ANN, RNN, recurrent graph network (2 types), PDE-driven graph network  
    & 42 river segments in Delaware River Basin 
\\
\cite{chen2020estimating} 
    & Evapo-transpiration (daily) 
    & DCBS 
    & LSTM, Temporal Convolution Network, ANN, RF, SVR, 7 different empirical models  
    & 16 weather stations in Northeast plain of China 
\\
\cite{choi2022utilization} 
    & Streamflow (daily) 
    & DCBS 
    & LSTM with different sets of inputs
    & 13 catchments in South Korea 
\\
\cite{corns2022deep} 
    & Stream water level (daily) 
    & DCBS 
    & LSTM ensembles
    & 20 catchments in Missouri 
\\
\cite{frame2021post} 
    & Streamflow (daily) 
    & DCBS, hybrid process-ML model 
    & LSTM, NWM reanalysis, LSTM+NWM hybrid
    & 531 catchments in US (CAMELS) 
\\
\cite{feng2021mitigating} 
    & Streamflow (daily) 
    & DCBS 
    & LSTM with different sets of encoded inputs
    & 671 catchments in US (CAMELS) 
\\
\cite{ghosh2022meta} 
    & Streamflow (daily) 
    & TL (meta transfer learning) 
    & LSTM, sequence autoencoder
    & 191 river segments in Delaware River Basin 
\\
\cite{jiang_improving_2020} 
    & Streamflow (daily) 
    & DCBS 
    & ANN, LSTM, KGML (custom network architecture)
    & 450 basins in US (CAMELS) 
\\
\cite{nogueira2022deep} 
    & Streamflow (monthly) 
    & DCBS 
    & LSTM, ANN, SMAP conceptual model
    & 25 catchments in Brazil 
\\
\cite{kalin2010predicting} 
    & 8 river water quality variables (daily) 
    & DCBS 
    & ANN with varying inputs
    & 18 monitoring locations in west Georgia, USA 
\\
\cite{koch2022long} 
    & Streamflow (daily) 
    & DCBS 
    & LSTM, DK process model  
    & 301 basins in Denmark 
\\
\cite{kratzert2019toward_ung} 
    & Streamflow (daily) 
    & DCBS 
    & LSTM, SAC-SMA process model, NWM reanalysis  
    & 531 basins in USA (CAMELS) 
\\
\cite{lee2020estimating} 
    & Maximum Streamflow (annual) 
    & DCBS, Hybrid process-ML model 
    & ANN, RF, RNN, SVR  
    & 64 catchments in South Korea 
\\
\cite{li2021improved} 
    & Daily Soil Moisture 
    & TL   
    & CNN, LSTM, ConvLSTM  
    & 3380 locations in China 
\\
\cite{ma_transferring_2021} 
    & Streamflow (daily) 
    & DCBS with TL 
    & LSTM   
    & 1389 cathments in US, China, UK, and Chile  
\\
\cite{muhebwa2021towards} 
    & Streamflow (daily) 
    & DCBS subgroup of entities broad-scale model 
    & LSTM (not directly compared)  
    & 5 classes of catchments in Canada 
\\
\cite{noori2020water} 
    & 3 water quality nutrient loads (monthly) 
    & DCBS, hybrid process-ML model 
    & ANN, SWAT process model, Hybrid SWAT+ANN
    & 29 monitoring locations in Georgia, USA 
\\
\cite{ouyang2021continental} 
    & Streamflow (daily) 
    & DCBS, subgroup of entities model 
    & LSTM 
    & 3557 basins in USA 
\\
\cite{potdar2021toward} 
    & Maximum streamflow (annual) 
    & DCBS 
    & XGB 
    & 3490 stream gauges in USA 
\\
\cite{rahmani2021deep} 
    & Stream temperature (daily) 
    & DCBS 
    & LSTM 
    & 455 basins in USA 
\\
\cite{rasheed2022advancing} 
    & Flood peaks (>90\% quantile streamflow) (daily) 
    & DCBS 
    & LSTM, RF, gradient boosting 
    & 670 catchments in USA (CAMELS) 
\\
\cite{razavi2016improving} 
    & Streamflow (daily) 
    & Hybrid Process-ML, subgroup of entities broad-scale model 
    & ANN, 2 process-based models (MAC-HBV and SAC-SMA) 
    & 90 watersheds in Ontario, Canada 
\\
\cite{singh2022streamflow} 
    & Streamflow (daily) 
    & TL 
    & SVR, XGB, SWAT process model 
    & 6 catchments in India 
\\
\cite{sun2021explore} 
    & Streamflow (daily) 
    & DCBS, broad-scale graph ML model   
    & 3 GNN architectures, LSTM 
    & 530 basins in USA (CAMELS) 
\\
\cite{tayal2022invertibility} 
    & Lake temperature at depth (daily) 
    & DCBS, broad-scale with encoding of site characteristics  
    & LSTM with varied encoder networks 
    & 450 lakes in Midwest USA 
\\
\cite{vaheddoost2023rainfall} 
    & Streamflow (daily) 
    & TL, hydrid process-ML  
    & RF, MARS, DAR process model 
    & 10 gauging stations on the Coruh River in T\"{u}rkiye 
\\
\cite{wang2022exploring} 
    & Snow water equivalent (daily) 
    & DCBS, TL  
    & LSTM, SN17 process model
    & 30,000 4km resolution pixels across USA 
\\
\cite{weierbach2022stream} 
    & Stream temperature (monthly) 
    & DCBS  
    & XGB, MLR, SVR 
    & 93 monitoring stations in Mid-Atlantic and Pacific Northwest USA 
\\
\cite{white2017predicting}
    & Stream temperature (monthly) 
    & DCBS  
    & RF, MLR, BCM process model 
    & 69 basins in California, USA 
\\
\cite{willard_predicting_2021} 
    & Lake temperature at depth (daily) 
    & TL (meta TL), KGML (informed loss, simulation pre-train) 
    & LSTM, GLM process model 
    & 450 lakes in Midwest USA 
\\
\cite{willard_daily_2022} 
    & Lake surface temperature (daily) 
    & DCBS  
    & LSTM, ERA5 reanalysis, linear model 
    & 185,549 lakes in USA 
\\
\cite{willard2023machine} 
    & Stream temperature (daily) 
    & DCBS, TL (meta TL)  
    & LSTM, XGBoost, linear model 
    & 1367 stream sites in USA 
\\
\cite{xiong2022predicting} 
    & Riverine nitrogen export (daily) 
    & DCBS, TL  
    & LSTM 
    & 7 watersheds across the world
\\
\cite{xu2023deep}
    & Streamflow (daily) 
    & TL 
    & Transformer, ANN, LSTM, TOPMODEL (process-based) 
    & 8 basins in China 
\\
\cite{yin_rainfall-runoff_2021} 
    & Streamflow (daily) 
    & DCBS 
    & LSTM with attribute-weighting module and multi-head-attention module 
    & 531 basins in USA (CAMELS) 
\\
\cite{yin2023runoff} 
    & Streamflow (daily) 
    & DCBS 
    & LSTM, Modified transformer with input transformation and custom position embedding 
    & 241 basins in USA (CAMELS) 
\\
\cite{zhi2021hydrometeorology} 
    & Riverine dissolved oxygen (daily) 
    & DCBS 
    & LSTM  
    & 236 watersheds in USA (CAMELS) 
\\
\cite{zhou2022flooddan} 
    & Flood forecasting (6 hour scale) 
    & DCBS, TL 
    & Unsupervised Domain Adaptation with LSTM, TCN, and GRU  
    & 2 watersheds in China and Taiwan 
\end{longtable}
\section{Summary and Discussion}
\label{sect:discussion}
We see that many variations of the three classes of ML methodologies discussed in Section \ref{sect:frameworks} have been used for \change{PUBs}{predictions in unmonitored sites} (Table \ref{table:lit}). So far, entity-aware broad-scale modeling through direct concatenation of features remains the dominant approach for hydrological applications. It remains to be seen how these different methods stack up against each other when predicting different environmental variables since most of current studies are on streamflow prediction. The evidence so far suggests that combining data from heterogeneous regions when available should be strongly considered. In Section \ref{subsect:broad_scale}, we saw many applications in which using \textit{all} available data across heterogeneous sites was the preferred method for training ML models as opposed to fitting to individual or a subset of sites. Many recent studies continue the traditional practice of developing unsupervised, process-based, and data-driven functional similarity metrics and homogeneity criteria when selecting either specific sites or subgroups of sites to build models on to be transferred to unmonitored sites. Notably, some of these works show models built on subgroups of sites outperform models using all available sites. Additionally, the results from \citet{frame2021post} suggest that using a broad-scale entity-aware ML model combining data from all regions is preferable to two different hybrid process-ML frameworks that harness a well-known process-based model in the NWM. Similarly, the results from \citet{fang2022data} suggest that deep learning models perform better when fed a diverse training dataset spanning multiple regions as opposed to homogeneous dataset on a single region even when the homogeneous data is more relevant to the testing dataset and the training datasets are the same size. This can likely be attributed to the known vulnerability of ML models that perform better when fed data from a diverse or slightly perturbed dataset (e.g. from adversarial perturbations), where they are able to learn the distinctions in underlying processes (see \cite{hao2022adversarially} for an example in hydrology).

It is also clear that the LSTM model remains by far the most prevalent neural network architecture for water resources time series prediction due to its natural ability to model sequences, its memory structure, and its ability to capture cumulative system status. We see that 30 of the 40 reviewed studies in Table \ref{table:lit} use LSTM. This aligns with existing knowledge and studies that have consistently found that LSTM is better suited for environmental time series prediction than traditional architectures without explicit cell memory \citep{zhang2018developing,fan2020comparison}. Even though we see the traditional ANN sometimes perform nearly as well or better  \citep{chen2020estimating,nogueira2022deep}, the LSTM has the advantage of not having to consider time-delayed inputs, which is a critical hyperparameter, due to its recurrent structure already incorporating many previous timesteps. We find that other neural network architectures suitable for temporal data like transformers \citep{vaswani_attention_2017} and temporal convolution networks (TCN) \citep{lea2017temporal} are not used much for unmonitored water resources applications compared to other disciplines doing sequential modeling such as natural language processing and bioinformatic sequence analysis where these methods have largely replaced LSTM. This is likely due to their recent development compared to LSTM and also possibly due to their lack of inclusion in major deep learning software packages like Pytorch and Keras. \add{One recent study by \protect\cite{yin2023runoff} seems to suggest that transformers outperform LSTM for rainfall-runoff prediction in the United States, but still the vast majority of transformer applications in hydrology are in the context of prediction in monitored sites \protect\citep{liu2023probing,liu2022improved,wei2023evaluate,xu2023transformer,wang2023temporal,yin2022rr}. How transformers fair in predicting in the unmonitored scenario will be an important research direction because results have been mixed in the monitored scenario when compared to LSTM with some showing improvement (e.g. \protect\citep{yin2022rr,liu2022improved}) and some not (e.g. \protect\citep{liu2023probing,wei2023evaluate}}). 

We also find that most studies are focused on daily predictions, although a few studies predict at a monthly, annual, or hourly time scales based on desired output resolutions, data availability  computational efficiency, or available computational power. For instance, monthly predictions may be desirable over daily due to the ability to use more interpretable, computationally efficient bootstrap ensembles, and easy-to-implement classical ML models \citep{weierbach2022stream}. Increased computational efficiency can also enable running a large number (e.g. millions) of model trainings or evaluations for parameter sensitivity or uncertainty analysis. 

Spatially, the majority of studies cover the United States at 27 out of 40 studies. 15 of these span the entire conterminous United States, while 10 are specific regions. The remaining studies are specific to certain countries and span Asia (7 studies), South America (1 study), Europe (3 studies), other North America (2), and two studies cover multiple continents.  The strong focus on the United States can be due to its large land area with rivers alongside the economic capability to have advanced monitoring stations where data are freely available for study worldwide. 

We also see the prevalence of the CAMELS dataset being used in streamflow studies; it is used in 9 out of the 40 studies in Table \ref{table:lit}. CAMELS serves as transformative continental-scale benchmark dataset for data-driven catchment science with its combined high quality streamflow measurements spanning 671 catchments, climate forcing data, and catchment characteristics like land cover and topography. However, we note that it is limited to "unimpaired" catchments that are not influenced by human management via dams. In addition to dam managed catchments, catchments close to and within urban areas excluded from CAMELS  are more likely to be impacted by roadways or other infrastructure. There are over 800,000 dammed reservoirs affecting rivers around the world, including over 90,000 in the United States \citep{international_rivers_2007,national_dams_2020}. The effect of dammed reservoirs on downstream temperature is also complicated by variable human-managed depth releases and changing demands for water and energy that affect decision making \citep{risley2010effects}.  These limitations may hamper the ability of current models to extrapolate to real-world scenarios where many catchments of high economic and societal value are either strongly human-impacted or data-sparse. 

\subsection{Open Questions for Further Research}
\label{sect:open_questions}
Though the works  reviewed in this survey encompass many techniques and applications, there are still many open issues to be addressed as the water resources scientific community increasingly adopts ML approaches for \change{PUB}{unmonitored} prediction. Here we highlight questions for further research that are widely applicable and agnostic to any specific target environmental variable, and should be considered as the field moves forward.

\subsubsection{Is more data always better?}
\label{subsec:future_optimal_training}
We have seen that deep learning models in particular benefit from large datasets of heterogeneous entities, challenging the longstanding notion transferring models between systems requires that they must be functionally similar \citep{razavi2013streamflow,guo2021regionalization}. Further research is needed to develop robust frameworks to discern how many sites need to be selected for training, what similarity needs to be leveraged to do so, and if excluding sites or regions can benefit broad-scale ML models when given different environmental variable prediction tasks. We hypothesize that excluding sites deemed dissimilar often limits the spectrum of hydrological heterogeneity, and the utilization of all available stream sites ensures a more comprehensive understanding of the system by allowing the model to learn from wide range of hydrological behaviors to more effectively generalize to unseen scenarios. This is supported by work in streamflow modeling that has explicitly analyzed the effect of merging data from heterogenous entities on prediction performance. \citep{fang2022data} is a great example demonstrating one step in deciding between using all available data versus a subset of functionally similar entities. Furthermore in stream temperature modeling, \citet{willard2023machine} also find using more data is beneficial for nearly all regions in the US, and both regional modeling and single-site modeling can benefit from being pre-trained on all available data. Moving forward we expect the use of a maximal amount of training data to be the default approach, especially given the advancements in computational power and hydrology having comparatively smaller datasets than other fields  where deep learning models are also commonly used like natural language processing, social media, and e-commerce.

\subsubsection{How do we select optimal training data and input features for prediction?}
\label{subsec:future_optimal_feature_and_data}
If it is not feasible or desirable to use all available data, this further begs the question of how to optimally select functionally similar entities to construct a training dataset to minimize target site prediction error. Many approaches exist to derive an unsupervised similarity between sites including using network science \citep{ciulla2022interpretable}, using metalearning to select training data (e.g. active learning-based data selection \citep{al2021data}), or comparing existing expert-derived metrics like hydrological signatures \citep{mcmillan2021review}. There are also methods to combine training for large-scale entity-aware modeling while also specifying a target region or class of similar sites exist (further explained in Section \ref{subsec:future_specificity_of_place}), and this is another example of where functional similarity could be applied. 

Approaches also exist to use ML frameworks like neural networks to develop the similarity encodings themselves, which could be used to select subgroups of sites. \cite{kratzert2019towards_og} demonstrate a custom LSTM architecture that delineates static and dynamic inputs, feeding the former to the LSTM input gate and the latter to the remaining gates. The idea is to use the input gate nodes to encode the functional similarity between stream gauge locations based on the site characteristics alone, and they show this to reveal interpretable hydrological similarity that aligns with existing hydrological knowledge. This framework as-is will not exclude any sites directly, but still offers insight into the usefulness of embedded functional similarity.  We also see the static feature encoding from Section \ref{subsubsec:encoding}, differing from the previously mentioned method by using a separate ANN for static features as opposed to different gates in the same LSTM. Future research in developing these similarity encodings can also extend into adversarial-based ML methods that could discern valuable training entities.  

Numerous other factors can be considered in training dataset construction when deciding whether to include entities other than functional similarity as well. First, the training data should be representative of all types of entities relevant to the prediction tasks, and not too biased towards a particular region or type of site which can correspondingly bias results. When building a model to transfer to a particular set of unmonitored sites, it must be considered whether the training data is representative of those target sites because environmental monitoring paradigms from the past that make up the dataset may not be in line with current priorities. Another consideration is the quality of data, where some sites may have higher quality of data than other sites which may have some highly uncertain characteristics. In cases like these, uncertainty quantification methods can be used to increase the reliability of predictions \citep{abdar2021review}, or different weighting can be assigned to different entities based on uncertainty metrics or what the training dataset needs to be representative. It has also been shown that assigning a vector of random values as a surrogate for catchment physical descriptors can be sufficient in certain applications \citep{li2022regionalization}. 

Furthermore, hydrologic prediction problems often contain a vast array of possible input features and input feature combinations spanning both dynamic forcing data like daily meteorology and static site characteristics. The process of feature selection aims to find the optimal subset of input features that (1) contains sufficient predictive information for an accuracy model, and (2) excludes redundant and uninformative features for better computational efficiency and model interpretability \citep{dhal2022comprehensive}. Notably, the majority of works reviewed in this study do not incorporate data-driven or statistical feature selection methods, and instead explicitly or presumably rely on expert domain knowledge to select inputs. This contrasts with many disciplines applying ML regression where feature selection is normalized and often deemed necessary (e.g. medical imaging  \citep{remeseiro2019review}, multi-view learning \citep{zhang2019feature}, finance \citep{khan2020stock}).  However, modern large-sample hydrology datasets offer a wealth of watershed, catchment, and individual site-specific characteristics and metrics that could serve as an opportunity to apply feature selection methods. For instance, the StreamCat dataset \citep{hill2016stream} contains over 600 metrics for ~2.65 million stream segments across the United, and the Caravan dataset \citep{kratzert2023caravan} contains ~70 catchment attributes for 6830 catchments across the world. 

Feature selection methods span three primary categories. \textit{Filter} feature selection methods rank variables based on their statistical properties alongside the target variable without considering the ML model itself. Popular filter techniques base rankings on correlation coefficients, mutual information, and information gain per feature. These methods have low computational cost compared to other methods, however they contain the drawback of not considering the interaction with the underlying ML model's performance. \textit{Wrapper}  feature selection methods on the other hand, assess the quality of variables by evaluating the performance of a specific ML model using a subset of features. Common wrapper methods include forward selection, backward elimination, boruta \citep{kursa2010feature}, and recursive feature elimination. These methods have the advantage of considering the interaction between variables and the model's performance, however they are more computationally expensive due to an often large number of model trainings and evaluations, especially for datasets which a large number of candidate input features. \textit{Embedded} (or intrinsic) feature selection methods are models that automatically already perform feature selection during training. Techniques like Lasso (Least Absolute Shrinkage and Selection Operator) and Elastic Net regularization automatically penalize the coefficients of irrelevant features during training, encouraging their removal. Additionally, random forest and similar decision tree methods also contain embedded feature selection as they will not include irrelevant features in the decision trees. 

The size and dimensionality of the hydrological dataset play a significant role in selecting a feature selection method. For large datasets with hundreds or thousands of possible features, filter methods can provide a computationally efficient initial screening. In contrast, wrapper methods such as recursive feature elimination or forward feature selection, are suitable for smaller datasets with fewer predictors, as they explicitly consider the regression model's performance. As hydrological modeling increasingly incorporates deep learning, the use of embedded methods may not be desirable since these methods generally exist among classical ML models. There is room for the hydrology community to develop standard processes to select optimal features for a given target variable and set of modeling sites. Furthermore, there needs to be methods to combine datasets where, for example, site-specific characteristics that need to be considered in a feature selection framework exist across multiple data sources.

\subsubsection{How should site characteristics be used in machine learning models for unmonitored prediction?} 
\label{subsec:future_specificity_of_place}

We have seen that generalization of ML models to unmonitored sites requires the availability of site characteristics \citep{kratzert2019toward_ung,xie2022estimating}, but that the science about how to use them is uncertain. The entity-aware models listed in this study tend to exhibit performance increases when such characteristics are included. For example, \cite{rasheed2022advancing} find site characteristics like soil porosity, forest fraction, and potential evapotranspiration all exhibit significant importance for flood peak prediction, and \cite{xie2022estimating} find that the combined catchment characteristics make up 20 percent of the total feature importances for a continental-scale baseflow prediction model. However, the result from \citet{li2022regionalization} showing random values substituted for site characteristics still improves performance in the temporal prediction scenario needs to be further investigated and compared in other applications. Many methods in this survey use site characteristics in different ways, and an open question remains of how to best add site characteristics to an ML model in a given task. 

Throughout this review, we see several ways to incorporate site characteristics into ML model architecture and frameworks. The most common way is in an entity-aware model using concatenated input features as seen in Section \ref{subsubsec:direct_concat}, presumably based on landmark results from the streamflow modeling community. However, it has also been demonstrated that using a graph neural network approach using these site characteristics to determine similarity between sites can slightly outperform the concatenated input approach \citep{sun2021explore}. Site characteristics have also been used to build and predict with a metamodel the performance of different local models to be transferred to an unmonitored site \citep{willard_predicting_2021,ghosh2022meta}. Other works mentioned in section \ref{subsubsec:encoding} demonstrate the effectiveness of learning ML-based encodings of site characteristics as opposed to using them as-is \citep{tayal2022invertibility,ghosh2022meta}. However, these approaches have not been tested against the concatenated input entity-aware approach commonly seen in other works which is needed to assess their role in \remove{PUB} modeling \add{unmonitored sites}. 

Furthermore, water management stakeholders, decision-makers,  and forecasters often seek to prioritize specific individual locations which are unmonitored but the site characteristics are known. Many of the broad-scale approaches \remove{to PUBs} mentioned in this survey are built without any knowledge of the specific testing sites they are going to be applied to. While training without any knowledge of the testing data is a common practice in supervised machine learning, \remove{PUBs} efforts \add{to predict in unmonitored sites} may benefit from including information on specific test sites during training. For example, characteristics from the test sites are used in the meta transfer learning framework described in Section \ref{subsec:transfer_learn} to select source models to apply to the target or test system. Surveys on transfer learning \citep{niu2020decade,pan_survey_2010} have described this distinction as the difference between \textit{inductive} transfer learning, where the goal is to find generalizable rules that apply to completely unseen data, with \textit{transductive} transfer learning, where the input data to the target or test system is known and can be used in the transfer learning framework. Transductive transfer learning methods like meta transfer learning have been proposed, but there is a lack of transductive methods that can harness the power of the highly successful entity-aware broad-scale models. In the same way that transfer learning has facilitated the pre-training of ML models in hydrology on data-rich watersheds to be transferred and fine tuned efficiently with little data in a new watershed, for example in flood prediction \citep{kimura2019convolutional}, we imagine there could be ways to harness to benefits of large-scale entity-aware modeling and also fine tune those same models to a specific region or class of sites that have known site characteristics. For example, the entity-aware models using all available data described in Section \ref{subsect:broad_scale} could be fine tuned to specific relevant subgroups, or the individual source models described in transfer learning approaches in Section \ref{subsec:transfer_learn} could be pre-trained using all available data \citep{willard2023machine}. 

There is also the issue of the non-stationary nature of many site characteristics. These characteristics are typically derived from synthesized data products that treat them as static values such as the GAGES-II (Geospatial Attributes of Gages for Evaluating Streamflow \citep{falcone_gages-ii_2011} containing basin topography, climate, land cover, soil, and geology), StreamCat \citep{hill2016stream}, and the dataset in \cite{willard_data_2021} (lake characteristics like bathymetry, surface area, stratification indices, and water clarity estimates). Though this treatment of site characteristics as static is intuitive for properties that do not evolve quickly (e.g. geology),  in reality properties such as land cover, land use, or even climate are dynamic in nature and evolve at different time scales. This can affect prediction performance in cases where the dynamic nature of certain characteristics treated as static are vital to prediction. For example, land use is a key dynamic predictor for river water quality in areas undergoing urbanization \citep{yao2023land}, but is treated as static in most hydrological ML models. In lake temperature modeling, water clarity is treated as static in \cite{willard_predicting_2021} but realistically has a notable dynamic effect on water column temperatures \citep{rose2016climate}. Though this problem exists in both monitored and \change{PUB}{unmonitored} scenarios, characteristics are particularly important in \change{PUB}{unmonitored site} prediction since often that is the only knowledge available concerning a location. As data collection from environmental sensors continues to improve, this highlights a need for new geospatial datasets and methods to represent dynamic characteristics at multiple time points (e.g. National Land Cover Database \citep{homer2012national}.

\subsubsection{How can we leverage process understanding for prediction in unmonitored sites?}
\label{future_kgml}
The success of ML models achieving better prediction accuracy across many hydrological and water resources variables compared to process-based models has led to the question posed by \cite{nearing_what_2021} of, "What role will hydrological science play in the age of machine learning?". Given the relevant works reviewed in this study showing mixed results comparing KGML approaches using process understanding with domain-agnostic black box approaches, more research is required to address the role of domain knowledge in \remove{PUB} prediction \add{for unmonitored sites}. From Section \ref{subsect:kgml} we see that using graph neural networks has potential to encode spatial context relevant for \remove{PUB} prediction\add{s} and improving over existing methods, but also that hybrid models have not been as effective as domain-agnostic entity-aware LSTM counterparts. A key research direction will be finding which context is relevant to encode in graphs or other similarity or distance-based structures, whether that be spatial or based on expert domain knowledge. A preferable alternative to existing hybrid process-ML models may be the DPB models explained in Section \ref{subsubsec:dpb}, which exhibit many side benefits like being able to output accurate intermediate variables and demonstrating interpretability, but the performance achieved remains similar to existing process-agnostic models like the entity-aware LSTM models. There is potential to further research and develop these DPB approaches, for instance they stand to benefit from assimilating multiple data sources since they simulate numerous additional variables. 

KGML modeling techniques, like informed loss functions, informed model architecture, and hybrid modeling can be considered during method development. For example, knowledge-guided loss function terms can impose structure on the solution search space in the absence of labeled target data by forcing model output to conform to physical laws (e.g. conservation of energy or mass). Examples of successful implementations of knowledge-guided loss functions to improve temporal prediction include the conservation of energy-based term to predict lake temperature \citep{read_process-guided_2019}, power-scaling law-based term to predict lake phosphorous concentration \citep{hanson_predicting_2020}, and advection–dispersion equation-based terms to predict subsurface transport states \citep{he2020physics}. These results show that informed loss functions can improve physical realism of the predictions, reduce the data required for good prediction performance, and also improve generalization to out-of-sample scenarios.  Since loss function terms are generally calculated on the model output and do not require target variable data, they can easily be transferred from temporal predictions to the \change{PUB}{unmonitored prediction} scenario.

Knowledge-guided architecture can similarly make use of the domain-specific characteristics of the problem being solved to improve prediction and impose constraints on model prediction, but has not been applied in the \change{PUB}{unmonitored} scenario. As opposed to soft constraints as imposed by a loss function term, architectural modifications can impose hard constraints. Successful examples of modified neural network architectures for hydrological prediction include a modified LSTM with monotonicity constraints for lake temperatures at different depths \citep{daw_physics-aware_2019},  mass-conserving modified LSTMs for streamflow prediction \citep{hoedt_mc-lstm_2021}, and an LSTM architecture that includes auxiliary intermediate processes that connect weather drivers to streamflow \cite{khandelwal_physics_2020}. Many hydrological prediction tasks involve governing equations such as conservation laws or equations of state that could be leveraged in similar ways to improve ML performance in unmonitored sites. 


We also see from Section \ref{subsect:kgml} that hybrid process and ML models are also a tool to consider for \change{PUBs}{ungauged and unmonitored prediction}. However, comparisons between different types of hybrid models are not commonly seen, as most studies we noted tend to use only one method. However, different types should be considered based on the context of the task. For example, if multiple process-based models are available then multi-model ensemble or using multiple process-based outputs as inputs to an ML model can be considered. Or, if part of the physical process is well-known and modeled compared to more uncertain components, researchers can consider replacing only part of the process-based model with an ML model component.


\subsubsection{How do we perform uncertainty quantification for predictions in unmonitored sites?}
Uncertainties in ML \remove{PUB} efforts \add{for prediction in unmonitored sites} can arise from various sources, including model structure and input data quality. Through uncertainty quantification (UQ) techniques, decision-makers can understand the limitations of the predictions and make informed decisions. UQ also enables model refinement, identification of data gaps, and prioritization of monitoring efforts in ungauged basins. Various techniques exist in UQ for ML \citep{abdar2021review}, including Bayesian deep learning \citep{wang2020survey}, dropout-based methods \citep{gal2016dropout}, Gaussian processes, and ensemble techniques. The concept of Bayesian deep learning is to incorporate prior knowledge and uncertainty by defining a full probability distribution for neural network parameters as opposed to a point estimate, which allows for the estimation of posterior distributions. These posterior distributions capture the uncertainty in the predictions and can be used to generate probabilistic forecasts in time series modeling. Gaussian processes similarly do Bayesian inference but over a particular function rather a deep neural network, and dropout methods approximate Bayesian ML by using a common regularization technique to randomly set a fraction of the parameters to zero, effectively "dropping them out" for a particular forward pass. This allows for the creation of an ensemble of models from a single model.

Using ensembles of models for prediction is a longstanding technique in hydrology that spans both process-based models \citep{thielen2008aims,troin2021generating} and more recently ML models \citep{zounemat2021ensemble}. Ensemble learning is a general meta approach to model building that combines the predictions from multiple models for both UQ and better predictive performance. In traditional water resources prediction, ideally, models in the ensemble will differ with respect to either meteorological input dataset (e.g. \cite{he2009tracking}), process-based model parameters (e.g. \cite{seibert2009gauging}) or multiple process-based model structures (e.g. \cite{moore2021lakeensemblr}). Different types of techniques are seen across ensemble learning more generally in the ML community, with common techniques such as (1) bagging, where many models are fit on different samples of the same dataset and averaging the predictions, (2) stacking, where different models types are fit on the same data and a separate model is used to learn how to combine the predictions, and (3) boosting, where ensemble members are added sequentially to correct the predictions made by previous models. Some of the main advantages of model ensembles in both cases is that the uncertainty in the predictions can be easily estimated and predictions can become more robust, leading them to be ubiquitous within many forecasting disciplines. Diversity in models is key, as model skill generally improves more from model diversity rather than from a larger ensemble \citep{delsole2014skill}.

There are key differences in ensemble techniques in process-based modeling versus ML. For instance, expert-calibrated parameters have very specific meanings in process-based models whereas the analogous parameters in ML (usually known as weights) are more abstract and characteristic of a black box. When tweaking parameters between models to assemble an ensemble, physical realism is important in the process-based model case. Parameterization has a rich history in process-based models and the work can be very domain-specific, whereas ML ensemble techniques are often done using existing code libraries through a domain-agnostic process. Furthermore, ML ensemble techniques usually do not modify input datasets, though they could through adding noise \citep{brownlee2018better} or by using different data products (e.g. for meteorology). 

We see most ML applications reviewed in this work do not attempt to use UQ techniques even though the few that do, see positive results (e.g. the use of ensembles for stream temperature \citep{weierbach2022stream} , streamflow \citep{feng2021mitigating}, and water level \citep{corns2022deep}). A recent survey by \cite{zounemat2021ensemble} finds that ensemble ML strategies demonstrate "absolute superiority" compared regular (individual) ML model learning in hydrology, and this result has also been seen in the machine learning community more generally for neural networks \citep{hansen1990neural}. Many opportunities exist to develop ensemble frameworks in water resources prediction that harness numerous diverse ML models. In the same way that the hydrology community often uses ensembles of different process-based model structures, the many different architectures and hyperparameters in deep learning networks can achieve a similar diversity. Given the common entity-aware broad-scale modeling approach seen widely throughout this review, opportunity exists to use resampling techniques like bootstrap aggregation \citep{breiman1996bagging} to vary training data while maintaining broad coverage, as seen in \cite{weierbach2022stream} for stream temperature. Other ensemble methods like in \cite{feng2021mitigating} vary which site characteristics are used as inputs to LSTMs for streamflow prediction. 

\subsubsection{What is the role of explainable AI in predictions for unmonitored sites?}
\label{future_derive_knowledge}
Historically, the difference between ML methods and more process-based or mechanistic methods has been described as a tradeoff between "predictive performance" and "explainability" \citep{lipton2018mythos}. However, there has been a deluge of advances in recent years in the field of explainable AI (XAI) \citep{arrieta2020explainable} and applications of these are increasingly being seen in geosciences \citep{bacsaugaouglu2022review,mamalakis2023carefully}. For example, recent work has shown how XAI can help to calibrate model trust and provide meaningful post-hoc interpretations \citep{toms2020physically}, identify how to fine-tune poor performing models \citep{ebert2020evaluation}, and also accelerate scientific discovery \citep{mamalakis2022investigating}. This has led to a change in the narrative of the performance and explainability tradeoff as calls are increasingly made for the water resources community to adopt ML as a complementary or primary avenue toward scientific discovery \citep{shen2018hess}.Though the majority of work using XAI in water resources time series prediction has been seen in the temporal prediction scenario (e.g. \cite{lees2022hydrological,kratzert2019neuralhydrology}), analysis of how ML models are able to learn and transfer hydrologic understanding for \remove{PUB} predictions \add{in unmonitored sites} can help address one of the most fundamental problems of "transferability" in hydrology. 

We find that many water resources studies still use classical ML models like random forest or XGBoost in part due to their ease of interpretability. Initial investigations of interpretability of deep learning frameworks have mostly addressed simple questions like feature attribution and sensitivity (e.g. \cite{sun2021explore,potdar2021toward}). The concept of DPB models discussed in Section \ref{subsubsec:dpb} shows potential to take this further and make an end-to-end interpretable model mimicking environmental processes but with the trainability and flexibility of deep neural networks. DPB models can provide more extensive interpretability compared to simpler feature attribution methods by being able to represent intermediate process variables explicitly in the neural network with the capability of extracting their relationship to the inputs and outputs. 

Future work on XAI for \change{PUBs}{unmonitored site prediction} can pose research questions in directions that harness the existing highly successful ML models to both refine theoretical underpinnings and add to the current hydrologic or other process understandings surrounding regionalizations to unmonitored sites. For example, methods like layerwise relevance propogation, integrated gradients, or Shapley additive explanations (SHAP) \citep{molnar2020interpretable} could be used to explore causations and attributions of observed variability in situations where ML predicts more accurately than existing process-based regionalization approaches. Both temporal and spatial attributes can be considered, for example when using methods like SHAP with LSTM the attributions of any inputs along the sequence length can be used to see how far back in time the LSTM is using its memory to perform predictions, or in GNNs to see where in space the knowledge is being drawn for prediction \citep{ying2019gnnexplainer}.

\section{Conclusion}
The use of ML for unmonitored environmental variable prediction is an important research topic in hydrology and water resources engineering, especially given the urgent need to monitor the effects of climate change and urbanization on our natural and man-made water systems. In this article, we review the latest methodological advances in ML for unmonitored prediction using entity-aware deep learning models, transfer learning, and knowledge-guided ML models. We summarize the patterns and extent of these different approaches and enumerate questions for future research. Addressing these questions sufficiently will likely require the training of interdisciplinary water resources ML scientists and also the fostering of interdisciplinary collaborations between ML and domain scientists. As the field of ML for water resources progresses, we see many of these open questions can also augment domain science understanding in addition to improving prediction performance and advancing ML science. We hope this survey can provide researchers with the state-of-the-art knowledge of ML for unmonitored prediction, offer opportunity for cross-fertilization between ML practitioners and domain scientists, and provide guidelines for the future.
\section*{Acronyms/Abbreviations}
\begin{tabular}{ll}
AI & Artificial Intelligence \\
ANN & Artificial neural network (feed forward) \\
CAMELS & Catchment Attributes and Meteorology for Large-sample Studies \\
DCBS & Direct concatenation broad-scale \\
DPB & Differentiable process-based \\
GNN & Graph neural network \\
GRU & Gated recurrent unit \\
KGML & Knowledge-guided machine learning \\
LSTM & Long short-term memory \\
MARS & Multi-adaptive regression splines \\
ML & Machine learning \\
NWM & National Water Model \\
RF & Random forest \\
XGB/XGBoost & Extreme gradient boosting \\
SHAP & SHapley Additive exPlanations \\
SVR & Support vector regression \\
TCN & Temporal convolutional network \\
XAI & eXplainable artificial intelligence \\
\end{tabular}

\begin{Backmatter}

\paragraph{Acknowledgments}
We are grateful for the editorial assistance of Somya Sharma, Kelly Lindsay, and Rahul Ghosh. \add{We also acknowledge the helpful comments from the anonymous reviewers, which helped improve this manuscript.}

\paragraph{Funding Statement}
This research is funded, in part, by NSF grants numbers 2313174, 2147195, 2239175, 2316305, 1934721 (HDR program), NSF LEAP Science and Technology Center award \#2019625, and National AI Research Institutes Competitive Award no. 2023-67021-39829. Additional support was provided by the U.S. Department of Energy, Office of Science, Biological and Environmental Research Program for the iNAIADS DOE Early Career Award under contract no. DE-AC02-05CH11231. This research used resources of the National Energy Research Scientific Computing Center (NERSC), a U.S. Department of Energy Office of Science User Facility located at Lawrence Berkeley National Laboratory, operated under Contract No. DE-AC02-05CH11231 under the NESAP for Learning program. The U.S. Government retains, and the publisher, by accepting the article for publication, acknowledges, that the U.S. Government retains a non-exclusive, paid-up, irrevocable, world-wide license to publish or reproduce the published form of this manuscript, or allow others to do so, for U.S. Government purposes.

\paragraph{Competing Interests}
Vipin Kumar is on the advisory board for the Environmental Data Science journal. 

\paragraph{Data Availability Statement}
Data sharing is not applicable to this article as no new data were created or analyzed in this study. 

\paragraph{Ethical Standards}
The research meets all ethical guidelines, including adherence to the legal requirements of the study country.

\paragraph{Author Contributions}
\textbf{Jared Willard}: Writing – Original Draft Preparation (lead); Conceptualization (equal); Data (literature) Curation (lead); Investigation (equal); Methodology (equal). \textbf{Charuleka Varadharajan}: Project Administration (supporting); Writing – Review \& Editing (equal); Supervision (\change{supporting}{equal})\add{; Funding Acquisition (equal)}. \textbf{Xiaowei Jia}: Conceptualization (supporting); Writing – Review \& Editing (supporting). \textbf{Vipin Kumar}: Conceptualization (equal); Investigation (equal); Project Administration (lead); Writing – Review \& Editing (equal); Supervision (lead); Methodology (equal); Funding Acquisition (equal).

\printbibliography

\end{Backmatter}

\end{document}